\documentclass[pdflatex,sn-mathphys-num]{sn-jnl}% Math and Physical Sciences Numbered Reference Style
\usepackage{graphicx}%
\usepackage{multirow}%
\usepackage{amsmath,amssymb,amsfonts}%
\usepackage{amsthm}%
\usepackage{mathrsfs}%
\usepackage[title]{appendix}%
\usepackage{xcolor}%
\usepackage{textcomp}%
\usepackage{manyfoot}%
\usepackage{booktabs}%
\usepackage{algorithm}%
\usepackage{algorithmicx}%
\usepackage{algpseudocode}%
\usepackage{listings}%
\usepackage{subcaption}

\theoremstyle{thmstyleone}%
\theoremstyle{thmstyletwo}%

\theoremstyle{thmstylethree}%

\begin{document}

\title[Adaptive Rainfall Forecasting from Multiple Geographical Models Using Matrix Profile and Ensemble Learning]{Adaptive Rainfall Forecasting from Multiple Geographical Models Using Matrix Profile and Ensemble Learning}


\author[2]{\fnm{Dung T. Tran}}\email{dung.tt2@vinuni.edu.vn}
\author[1]{\fnm{Huyen Ngoc Tran}}\email{huyen.tn2414629@sis.hust.edu.vn}
\author[3]{\fnm{Hong Nguyen}}\email{hongn@usc.edu}
\author[1]{\fnm{Xuan-Vu Phan}}\email{vu.phanxuan@hust.edu.vn}
\author*[1]{\fnm{Nam-Phong Nguyen}}\email{phong.nguyennam@hust.edu.vn}

\affil[1]{\orgname{Ha Noi University of Science and Technology}, \orgaddress{\city{Ha Noi}, \country{Vietnam}}}

\affil[2]{\orgdiv{Center of Environmental Intelligence}, \orgname{VinUniversity}, \city{Hanoi}, \country{Vietnam}}

\affil[3]{\orgname{University of Southern California}, \orgaddress{\city{Los Angeles}, \country{United States}}}


\abstract{Rainfall forecasting in Vietnam is highly challenging due to its diverse climatic conditions and strong geographical variability across river basins, yet accurate and reliable forecasts are vital for flood management, hydropower operation, and disaster preparedness. In this work, we propose a Matrix Profile-based Weighted Ensemble (MPWE), a regime-switching framework that dynamically captures covariant dependencies among multiple geographical model forecasts while incorporating redundancy-aware weighting to balance contributions across models. We evaluate MPWE using rainfall forecasts from eight major basins in Vietnam, spanning five forecast horizons (1 hour and accumulated rainfall over 12, 24, 48, 72, and 84 hours). Experimental results show that MPWE consistently achieves lower mean and standard deviation of prediction errors compared to geographical models and ensemble baselines, demonstrating both improved accuracy and stability across basins and horizons.}

\maketitle

\section{Introduction}
\label{sec:Introduction}
    Accurate forecasts reduce the impacts of floods, droughts, and typhoons, which cause major economic and human losses on a global scale each year. With climate change intensifying extreme rainfall events, the need for robust and reliable prediction methods is more urgent than ever \cite{RN4,1160787}.

    Geographical models such as global circulation models (GCMs) and regional climate models (RCMs) forecast rainfall at different spatial and temporal scales by capturing large-scale atmospheric dynamics. However, they often suffer from systematic biases and uncertainties at local levels, tending to underestimate extreme precipitation and overestimate light rainfall, which undermines reliability in operational decision-making \cite{Giorgi2002-tt,chen2011uncertainty,maraun2016bias}. In Vietnam, these challenges are exacerbated by the tropical monsoon climate, frequent typhoons, and complex orography. Existing models often underestimate heavy monsoon rainfall, and sparse rainfall station networks, particularly in mountainous and rural areas, further limit predictive accuracy \cite{tuyet2019performance,nguyen2023performance,thanh2014biascorrection}.

    Ensemble learning methods have been developed to improve forecasts by combining the output of multiple models. While early approaches used simple averaging or linear regression, more advanced methods such as Bayesian model averaging (BMA) \cite{raftery2005using}, quantile regression averaging (QRA) \cite{koenker1978regression}, ensemble model output statistics (EMOS) \cite{gneiting2005calibrated}, and machine learning ensembles (e.g., random forests, gradient boosting) \cite{taillardat2016calibrated} have shown promise in capturing nonlinear relationships. However, most ensembles rely on static weighting or simplistic correlation structures, making them less effective under rapidly changing weather regimes or extreme events.

    Recent advances in time-series analysis, particularly the matrix profile, offer efficient tools for motif discovery, discord detection, and similarity search \cite{yeh2016matrix,zimmerman2019matrix}. Despite their success in capturing dynamic temporal dependencies, matrix profiles have not been integrated into the predictions of the rainfall ensemble. This gap limits current ensembles from adapting to non-stationary conditions and correcting forecasts dynamically.
    
    To overcome these limitations, we propose a hybrid approach that integrates matrix profile analysis with ensemble learning for rainfall prediction. The main contributions of this work are:
    \begin{itemize}
        \item We develop a \textbf{dynamic ensemble learning framework} that integrates matrix profile analysis with multiple geographical forecasts. This allows the model to adaptively capture temporal dependencies across forecasts, going beyond the static weighting schemes used in conventional ensembles.  
        \item We perform a \textbf{comprehensive experimental study} on rainfall prediction in Vietnam, benchmarking against a wide spectrum of baselines, including individual climate models, statistical ensembles (e.g., BMA, QRA, EMOS), and machine learning approaches (e.g., random forest, XGBoost).  
    \end{itemize}
    
\section{Related Work}
\label{sec:related_work}
    \subsection{Geographical Models for Rainfall Prediction in Vietnam.}
        Rainfall forecasting in Vietnam heavily relies on configurations of the \textit{Weather Research and Forecasting} (WRF) model \cite{skamarock2008wrf}, a numerical weather prediction system widely used for high-resolution atmospheric simulation. Formally, each configuration $M_j$ generates a rainfall forecast $\hat{y}^{(j)}_t$ for time $t$, while the true observation is $y_t$. The prediction error can be expressed as $e^{(j)}_t = y_t - \hat{y}^{(j)}_t$. Different WRF setups such as \texttt{COMS}, \texttt{GFS}, \texttt{MITSUISHI2011\_D02} \cite{Mitsuishi2011}, \texttt{LING3\_D02}, \texttt{LINKF\_D02}, \texttt{LINBMJ\_D02}, \texttt{ETAKF\_D02}, \texttt{ETAG3\_D02}, and \texttt{ETABMJ\_D02}, which differ in physical schemes, grid resolution, and domain setup, leading to distinct patterns in $\hat{y}^{(j)}_t$ and systematic biases in $e^{(j)}_t$. 

        Studies have demonstrated the ability of WRF in simulating heavy rainfall and regional climate patterns in Vietnam \cite{Thang202201,raghavan2016regional}. However, the raw outputs $\hat{y}^{(j)}_t$ tend to underestimate extreme precipitation while overestimating light rainfall, leading to biased errors $e^{(j)}_t$ that reduce operational reliability.

    \subsection{Rainfall Prediction Approaches}
        Traditional rainfall prediction methods include statistical downscaling and dynamical downscaling from GCMs to RCMs. Statistical methods are computationally efficient but inadequate for capturing nonlinear atmospheric dynamics \cite{wilks2011statistical}, while dynamical methods are physically grounded but computationally expensive and biased in tropical regions such as Vietnam \cite{ngo2014climate,ngo2017performance}. Recent data-driven approaches, such as neural networks, SVMs, LSTMs, and CNNs, have shown promise \cite{gelete2023application,darji2019rainfall}. However, their dependence on large localized datasets poses challenges in developing countries where observations are sparse.    

    \subsection{Ensemble Forecasting in Climate Science}
        Ensemble forecasting mitigates the uncertainty of individual models by aggregating outputs. Methods such as Bayesian Model Averaging (BMA) \cite{raftery2005using}, regression-based ensembles \cite{krishnamurti1999improved}, and deep ensembles \cite{lakshminarayanan2017simple} generally outperform single-model forecasts. Hydrology has also benefited from ensemble methods for streamflow prediction, flood forecasting, and seasonal rainfall estimation \cite{krishnamurti1999improved}. Nonetheless, most ensembles apply static or error-based weights, overlooking temporal dependencies and covariances between model outputs, which can carry critical information about systematic biases and error propagation.

    \subsection{Matrix Profile for Time-Series Analysis}
        The Matrix Profile (MP) \cite{yeh2016matrix} is a scalable technique for time-series mining that supports motif discovery, anomaly detection, and similarity search. Its efficiency has led to widespread use in domains such as finance, healthcare, industrial IoT, and energy systems \cite{nilsson2023practical,de2020generalized}. Despite its success, MP has rarely been applied in climate science, and no prior studies have investigated its integration with ensemble rainfall prediction. This leaves open an opportunity to exploit MP’s ability to uncover dynamic dependencies across multiple forecasts.  

    \subsection{Research Gap}
        Although climate models, ensemble learning, and time-series mining have advanced rainfall prediction, several gaps remain. Forecasts from GCMs and RCMs remain uncertain in Vietnam due to monsoon dynamics and complex orography. Ensemble methods improve skill, but rely on static weights and ignore cross-model dependencies, allowing correlated errors to persist. MP offers tools to capture temporal dependencies, but has not been applied to rainfall ensemble forecasting. Addressing these limitations motivates our proposed framework, which integrates MP with ensemble learning to deliver dependency-aware rainfall forecasts tailored to Vietnam’s complex climatic conditions.

\section{Methodology}
\label{sec:methodoly}
    \begin{figure}
        \centering
        \includegraphics[width=\linewidth]{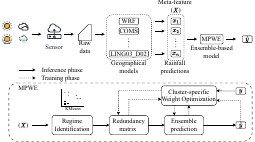}
        \caption{The overview framework of MPWE.}
        \label{fig:framework}
    \end{figure}
    \subsection{Overview of the proposed method}
        This work introduces MPWE, a novel ensemble framework that dynamically adjusts geographical model weights based on input characteristics. Unlike conventional ensembles that assign fixed weights to base models, our approach identifies latent regimes in the data and learns cluster-specific weights to optimize predictive performance. By incorporating redundancy-aware regularization, the combiner mitigates the influence of correlated models, resulting in a robust and interpretable ensemble. Figure \ref{fig:framework} illustrates the overall framework.
        \subsubsection{Regime Identification}
            Rainfall prediction is influenced by diverse sources of variability, including seasonal monsoon phases, geographical regions, and large-scale atmospheric drivers such as typhoons or El Niño events. This variability creates \emph{heterogeneity in the input space}, meaning that the statistical relationships between predictors (e.g., outputs of geographical models, meteorological variables) and the target rainfall can change across conditions. To account for this, we partition the data into distinct \emph{regimes}, where each regime represents a subset of samples characterized by similar climatic or meteorological patterns. Within each regime, the optimal combination of base models may differ, reflecting the fact that certain forecasts are more reliable under specific conditions.

            Let $\mathrm{X} \in \mathrm{R}^{n \times M}$ be the meta-feature matrix, where $n$ denotes the number of training samples and $M$ the number of geographical models. Each training sample corresponds to a time–location pair (e.g., a particular day at a specific station or grid cell) for which rainfall forecasts from all $M$ models are available. Thus, each row of $\mathrm{X}$ stores the $M$ model predictions for one sample, while $\mathrm{y} \in \mathrm{R}^{n}$ contains the corresponding observed rainfall values.

            We apply KMeans clustering to $\mathrm{X}$ to partition the samples into $K$ clusters. For each sample $i \in \{1, 2, \dots, n\}$, the cluster assignment $c_i \in \{1,2,\dots,K\}$ indicates which regime it belongs to. Collectively, the cluster assignments are: 
            
            \begin{equation}
                \{c_1, c_2,\dots,c_n\} \in \{1,2,\dots, K\}
            \end{equation}

            By clustering the samples into regimes, the combiner can learn cluster-specific weights, enabling it to respond to different patterns in the input predictions adaptively.
        \subsubsection{Redundancy-Aware Similarity Matrix}
            Base models, which are geographical models in this case, are often correlated, which can result in over-reliance on redundant predictions. To address this, we compute a redundancy matrix $\mathrm{S} \in \mathrm{R}^{M\times M}$ using MP analysis. The matrix profile captures temporal similarity between predictions, highlighting patterns that standard correlation may miss. Higher entries in $\mathrm{S}$ indicate greater redundancy between models. This redundancy matrix serves as a regularizer in the weight optimization step.
        \subsubsection{Cluster-Specific Weight Optimization}
            For each cluster, we solve a constrained quadratic optimization problem to obtain the optimal combination of base models:
            \begin{equation}
                \min_{w} \; \|y - X w\|^2 + \lambda w^\top S w,
            \end{equation}
                subject to:
            \begin{equation}
                w_i \in [0,1], \quad \sum_{i=1}^{M} w_i = 1,
            \end{equation}
            where $y$ is the ground-truth vector, $w \in \mathrm{R}^M$ is the weight vector for the cluster, and $\lambda$ controls redundancy regularization. 

            The first term minimizes prediction error, while the second discourages overweighting correlated models. Constraints ensure non-negative weights that sum to one, promoting interpretability. Optimization is performed using Sequential Least Squares Programming (SLSQP).
        \subsubsection{Adaptive Prediction}
            During inference, new samples are assigned to clusters using the trained KMeans centroids. For each new test sample $x_{\text{test}} \in \mathrm{R}^{M}$, the distance to each cluster centroid is computed, and the sample is assigned to the nearest cluster, denoted as $c_{\text{test}}$. The corresponding cluster-specific weight vector $w_{c_{\text{test}}}$, which was learned during training, is then retrieved and applied to compute the final prediction as
            \begin{equation}
                \hat{y}_{\text{test}} = \sum_{i=1}^{M} w_{c_{\text{test}}, i} \, x_{\text{test}, i}.
            \end{equation}
            This regime-dependent weighting allows the ensemble to emphasize the most informative models for each sample while reducing the influence of redundant base models, thereby improving predictive accuracy and robustness across diverse regimes.
    \subsection{Theoretical Justification of MPWE Superiority}
        Let $X \in \mathrm{R}^{n \times M}$ be the matrix of predictions from $M$ base models over $n$ samples, and let $y \in \mathrm{R}^n$ be the ground-truth vector. Traditional ensemble methods compute the final prediction as a weighted sum of base models with fixed weights $w \in \Delta^M$, where $\Delta^M = \{ w \in \mathrm{R}^M : w_i \ge 0, \sum_{i=1}^{M} w_i = 1 \}$:
        \begin{equation}
            \hat{y}_{\text{trad}} = X w.
        \end{equation}
        The corresponding mean squared error (MSE) loss is:
        \begin{equation}
            \mathcal{L}_{\text{trad}}(w) = \| y - X w \|^2.
        \end{equation}
        This fixed-weight approach does not adapt to heterogeneous input patterns and does not account for redundancy among correlated base models.
        \subsubsection{MPWE Formulation}
            To address these limitations, MPWE divides the data into $K$ clusters (latent regimes) based on the predictions $X$:
            \begin{equation}
                c_i = \text{cluster}(x_i), \quad i=1,\dots,n.
            \end{equation}
            For each cluster $k$, we optimize a cluster-specific weight vector $w_k \in \Delta^M$ with redundancy-aware regularization:
            \begin{equation}
                w_k = \arg\min_{w \in \Delta^M} \underbrace{\|y^{(k)} - X^{(k)} w\|^2}_{\text{Cluster MSE}} + \underbrace{\lambda w^\top S w}_{\text{Redundancy penalty}},
            \end{equation}
            where $X^{(k)}$ and $y^{(k)}$ denote the samples in cluster $k$, $S \in \mathrm{R}^{M \times M}$ captures correlations between base models, and $\lambda > 0$ controls the strength of redundancy regularization.
            
            The prediction for sample $i$ is then:
            \begin{equation}
                \hat{y}_i^{\text{MP}} = \sum_{j=1}^{M} w_{c_i,j} \, x_{i,j}.
            \end{equation}
        \subsubsection{Theoretical Superiority}
            Let $\mathcal{W}_{\text{trad}}$ and $\mathcal{W}_{\text{MP}}$ denote the feasible sets of weights for traditional ensembles and MPWE, respectively. Clearly,
            \begin{equation}
                \mathcal{W}_{\text{trad}} \subseteq \mathcal{W}_{\text{MP}},
            \end{equation}
            since any fixed weight $w \in \Delta^M$ is a special case of cluster-specific weights $w_k = w$ for all $k$.  

            Therefore, by optimizing cluster-specific weights and penalizing redundancy, the MSE of MPWE is guaranteed to be no larger than that of traditional ensembles:
            \begin{equation}
                \min_{\{w_k \in \Delta^M\}_{k=1}^{K}} 
                \sum_{k=1}^{K} \Big( \|y^{(k)} - X^{(k)} w_k\|^2 + \lambda w_k^\top S w_k \Big) 
                \le \min_{w \in \Delta^M} \|y - X w\|^2.
            \end{equation}
\section{Experiments}
\label{sec:experiment}
    \subsection{Datasets and Preprocessing}
        We use rainfall forecasts from eight Vietnamese basins (Ban Nhung, Ban Ve, Hua Na, Khanh Khe, Muong Hum, Song Chay 3, Song Chung, and Thac Xang) collected between June 2023 and July 2024. Hydropower dams in these basins issue forecasts twice daily (7:00 AM, 1:00 PM) for the next 84 hours, which serve as inputs to our ensemble models. All features were standardized using \texttt{StandardScaler}; no additional feature engineering or imputation was applied to focus on the ensemble methodology. 
    \subsection{Experimental Setup}
        MPWE hyperparameters were tuned with \texttt{GridSearchCV}, exploring number of clusters (\texttt{n\_clusters} $\in \{2,3,5\}$), Matrix Profile window size (\texttt{window} $\in \{5,10,20\}$), and redundancy regularization coefficient (\texttt{lam} $\in \{0.01,0.1,1.0\}$). The best configuration was selected based on cross-validated mean squared error (MSE). All experiments were run in Python 3.11 with Scikit-learn 1.2.2 on a standard workstation.
    \subsection{Baselines}
        We compare MPWE against geographical models and ensemble approaches, including Random Forest, XGBoost, Bayesian Model Averaging, regression-based ensembles, mean and median aggregation, and Quantile Regression Averaging. Each baseline was tuned using \texttt{GridSearchCV}, ensuring a fair comparison under optimal configurations.
    \subsection{Performance Evaluation}
        \begin{table}
            \centering
            \caption{Mean deviation of prediction errors for 2 basins: Ban Nhung and Ban Ve.}
            \label{tab:error_by_basin_1_mean}
            \begin{tabular}[t]{@{}l|cccccc@{}}
                \toprule
                \textbf{Basin / Method} & \textbf{1h} & \textbf{12h} & \textbf{24h} & \textbf{48h} & \textbf{72h} & \textbf{84h} \\
                \midrule
                \multicolumn{7}{c}{\textbf{Ban Nhung}} \\
                \botrule

                \hline
                COMS                          & 0.50              & 3.24                & 5.20             & 8.58             & 11.85           & 13.49 
                \\
                GFS                           & 0.48              & 2.88                & 4.51             & 7.14             & 9.53             & 10.64 \\
                MITSUISHI2011\_D02            & 0.78              & 6.28                & 11.92            & 23.51            & 35.22            & 41.07 \\
                LING3\_D02                    & 0.72              & 5.70                & 10.6             & 20.66            & 30.87            & 36.00 \\
                LINKF\_D02                    & 0.79              & 6.43                & 12.21            & 24.11            & 36.16            & 42.19 \\
                LINBMJ\_D02                   & 0.54              & 3.47                & 5.52             & 9.25             & 12.73            & 14.50 \\
                ETAKF\_D02                    & 0.67              & 5.01                & 9.14             & 17.62            & 26.52            & 30.59 \\
                ETAG3\_D02                    & 0.61              & 4.42                & 7.81             & 14.57            & 21.47            & 24.97 \\
                ETABMJ\_D02                   & 0.49              & 2.99                & 4.51             & 7.00             & 9.43             & 10.55 \\ 
                \hline
                Random Forest                 & 0.40              & 2.22                & 3.32             & 4.93             & 6.28             & 6.89 \\
                XGBoost                       & 0.38              & 2.05                & \underline{3.04} & \underline{4.35} & \underline{5.28} & \underline{5.65}
                \\
                Bayesian Model Averaging      & 0.38              & 2.27                & 3.32             & 4.92             & 6.21             & 6.74
                \\
                Regression-Based Method       & 0.40              & \underline{2.11}    & 3.05             & 4.4              & \underline{5.28} & 5.66
                \\
                Simple Mean Model             & 0.38              & 2.27                & 3.32             & 4.92             & 6.21             & 6.74 
                \\
                Simple Median Model           & 0.36              & 2.29                & 3.55             & 5.83             & 7.99             & 9.06
                \\
                Quantile Regression Averaging & \underline{0.30}  & 2.91                & 5.71             & 11.37            & 17.05            & 19.9
                \\ \hline
                \textbf{MPWE (ours)}          & \textbf{0.24}     & \textbf{1.92}       & \textbf{2.93}    & \textbf{4.26}    & \textbf{5.23}    & \textbf{5.45} \\
                \midrule
                \multicolumn{7}{c}{\textbf{Ban Ve}} \\
                \botrule
                COMS                          & 0.55              & 2.87                 & 4.30             & 6.52             & 8.50            & 9.51 \\
                GFS                           & 0.56              & 2.92                 & 4.32             & 6.48             & 8.27            & 9.21 \\
                MITSUISHI2011\_D03            & 1.08              & 7.60                 & 14.21            & 27.76            & 41.49           & 48.40 \\
                LINKF\_D03                    & 1.08              & 7.65                 & 14.18            & 27.66            & 41.39           & 48.29 \\
                LINBMJ\_D03                   & 0.80              & 4.45                 & 7.14             & 12.20            & 17.12           & 19.56 \\
                ETAKF\_D03                    & 0.87              & 5.08                 & 8.73             & 16.03            & 23.50           & 27.24 \\
                ETAG3\_D03                    & 0.67              & 3.34                 & 4.85             & 7.12             & 9.09            & 10.12 \\
                ETABMJ\_D03                   & 0.67              & 3.36                 & 4.96             & 7.45             & 9.62            & 10.65 \\ 
                \hline
                Random Forest                 & 0.56              & 2.74                 & 4.05            & 6.04             & 7.56            & 8.33  \\
                XGBoost              & 0.55              & \underline{2.62}     & \underline{3.88}      &\underline{5.69}    &\underline{7.07}    &\underline{7.75} \\
                Bayesian Model Averaging      & 0.54              & 3.19                 & 5.33             & 9.35             & 13.37           & 15.43 \\
                Regression-Based Method       & 0.58              & 2.76                 & 4.07             & 6.08             & 7.62            & 8.35 \\
                Simple Mean Model             & 0.54              & 3.19                 & 5.32             & 9.32             & 13.32           & 15.36 \\
                Simple Median Model           & 0.54              & 3.33                 & 5.74             & 10.43            & 15.20           & 17.62 \\
                Quantile Regression Averaging &\underline{0.45}   & 4.56                 & 9.08             & 18.16            & 27.22           & 31.75 \\ 
                \hline
                \textbf{MPWE (ours)}          & \textbf{0.50}     & \textbf{2.07}        & \textbf{3.78}    & \textbf{4.76}    & \textbf{6.87}   & \textbf{7.46} \\
                \botrule
                % Repeat rows
            \end{tabular}
        \end{table}

        \begin{table}
            \centering
            \caption{Standard deviation of prediction errors for 2 basins: Ban Nhung and Ban Ve.}
            \label{tab:error_by_basin_1_std}
            \begin{tabular}[t]{@{}l|cccccc@{}}
                \toprule
                \textbf{Basin / Method} & \textbf{1h} & \textbf{12h} & \textbf{24h} & \textbf{48h} & \textbf{72h} & \textbf{84h} \\
                \midrule
                \multicolumn{7}{c}{\textbf{Ban Nhung}} \\
                \botrule
                
                \hline
                COMS                          & 1.04              & 2.97                & 4.19             & 6.25             & 7.98           & 8.78 \\
                GFS                           & 0.95              & 2.65                & 3.79             & 5.52             & 6.88             & 7.56 \\
                MITSUISHI2011\_D02            & 1.30              & 4.37                & 6.69            & 9.74            & 11.92            & 12.86 \\
                LING3\_D02                    & 1.35              & 4.44                & 6.78             & 10.07            & 12.24            & 12.97 \\
                LINKF\_D02                    & 1.31              & 4.39                & 6.69            & 9.82            & 12.02            & 12.89 \\
                LINBMJ\_D02                   & 1.16              & 3.35                & 4.76             & 7.03             & 9.34            & 10.36 \\
                ETAKF\_D02                    & 1.13              & 3.74                & 5.71             & 8.66            & 10.73            & 11.51 \\
                ETAG3\_D02                    & 1.16              & 3.52                & 5.36             & 8.24            & 10.36            & 11.18 \\
                ETABMJ\_D02                   & 1.05              & 2.79                & 3.85             & 5.53             & 7.04             & 7.65 \\
                \hline
                Random Forest                 & \underline{0.67}              & \underline{1.66}                & \textbf{2.27}             & 3.29             & 4.14             & 4.51 \\
                XGBoost                       & \underline{0.67}              & 1.67                & \textbf{2.27} & \textbf{3.17} & \underline{3.92} & \underline{4.23} \\
                Bayesian Model Averaging      & 0.79              & 2.06                & 2.93             & 4.17             & 5.15             & 5.58 \\
                Regression-Based Method       & 0.68    & 1.70    & 2.37             & 3.22              & 3.97 & 4.27 \\
                Simple Mean Model             & 0.79              & 2.06                & 2.93             & 4.18             & 5.15             & 5.59 \\
                Simple Median Model           & 0.78              & 2.17                & 3.17             & 4.69             & 5.96            & 6.50 \\
                Quantile Regression Averaging & 0.78  & 2.66                & 3.85             & 5.50            & 6.63            & 7.11\\ \hline
                \textbf{MPWE (ours)}          & \textbf{0.66}     & \textbf{1.19}       & \underline{2.79}    & \underline{3.19}    & \textbf{3.90}    & \textbf{3.43} \\
                \hline
                \multicolumn{7}{c}{\textbf{Ban Ve}} \\
                \hline
               \botrule
                COMS                          & 0.97              & 2.77                & 3.98             & 5.74             & 7.40           & 8.19 \\
                GFS                           & 0.99              & 2.78                & 3.97             & 5.50             & 7.07             & 7.75 \\
                MITSUISHI2011\_D03            & 1.34              & 4.66                & 7.09            & 10.91            & 13.89            & 15.15 \\
                LINKF\_D03                    & 1.63              & 5.65                & 8.57            & 13.06            & 16.97            & 18.68 \\
                LINBMJ\_D03                   & 1.26              & 3.69                & 5.39             & 8.07             & 10.78            & 12.08 \\
                ETAKF\_D03                    & 1.13              & 3.45                & 5.34             & 8.60            & 11.40            & 12.73 \\
                ETAG3\_D03                    & 1.07              & 2.86                & 4.08             & 5.67            & 7.13            & 7.78 \\
                ETABMJ\_D03                   & 1.10              & 2.99                & 4.18             & 5.87             & 7.23             & 7.89 \\
                \hline
                Random Forest                 & 0.85              & \underline{2.37}                & \underline{3.28}             & \textbf{4.36}             & \underline{5.52}             & \underline{6.07} \\
                XGBoost                       & \underline{0.84}              & 2.39                & 3.34 & 4.49 & 5.65 & 6.24 \\
                Bayesian Model Averaging      & 1.00              & 3.11                & 4.64             & 7.13             & 9.45             & 10.55 \\
                Regression-Based Method       & 0.86    & 2.42    & 3.42             & 4.62              & 5.89 & 6.51 \\
                Simple Mean Model             & 1.00              & 3.11                & 4.64             & 7.13             & 9.45             & 10.55 \\
                Simple Median Model           & 1.01              & 3.23                & 4.85             & 7.52             & 9.89            & 11.03 \\
                Quantile Regression Averaging & 1.02  & 3.68                & 5.37             & 7.74            & 9.77            & 10.75\\ \hline
                \textbf{MPWE (ours)}          & \textbf{0.82}     & \textbf{2.24}       & \textbf{2.88}    & \underline{4.45}    & \textbf{4.63}    & \textbf{5.65} \\
                \hline
            \end{tabular}
        \end{table}
        Tables \ref{tab:error_by_basin_1_mean}, \ref{tab:error_by_basin_1_std}, \ref{tab:error_by_basin_2_mean}, \ref{tab:error_by_basin_2_std}, \ref{tab:error_by_basin_3_mean}, \ref{tab:error_by_basin_3_std}, \ref{tab:error_by_basin_4_mean} and \ref{tab:error_by_basin_4_std} present the comparison between our approach and the baselines, including geographical models and ensemble learning methods, across five forecasting scenarios (1-hour and accumulated rainfall at 12, 24, 48, 72, and 84 hours) over eight basins. Evaluation was based on the mean and standard deviation of prediction errors, where a good method should achieve both a low mean (high accuracy) and a low standard deviation (stable performance). By this criterion, our approach achieves the most accurate and reliable forecasts overall, consistently outperforming all baselines. XGBoost ranks as the second-best baseline, reflecting its strength in modeling nonlinear dependencies, although its error statistics remain consistently higher than ours. In the short-term (1-hour) scenario, Quantile Regression Averaging (QRA) shows competitive performance, demonstrating its effectiveness for uncertainty handling in immediate forecasts, though its performance deteriorates over longer horizons.  

        In contrast, geographical models with basin-specific configurations show large variability. While they sometimes perform well in individual basins due to targeted calibration, their lack of generalization results in higher errors when averaged across all basins. Ensemble-based methods improve on these models by balancing performance across basins, yet they remain specialized to certain conditions. It is also worth noting that Vietnam has highly diverse climatic conditions, ranging from tropical monsoon to subtropical highland regions, which introduces substantial spatial heterogeneity and makes rainfall forecasting particularly challenging. The superior performance of our approach across horizons and basins highlights its robustness and adaptability, effectively combining localized basin dynamics with broader spatiotemporal dependencies to address these climatic complexities.
        
        \begin{figure}
            \centering
            % First column
            \begin{subfigure}[t]{0.48\linewidth}
                \centering
                \includegraphics[width=\linewidth]{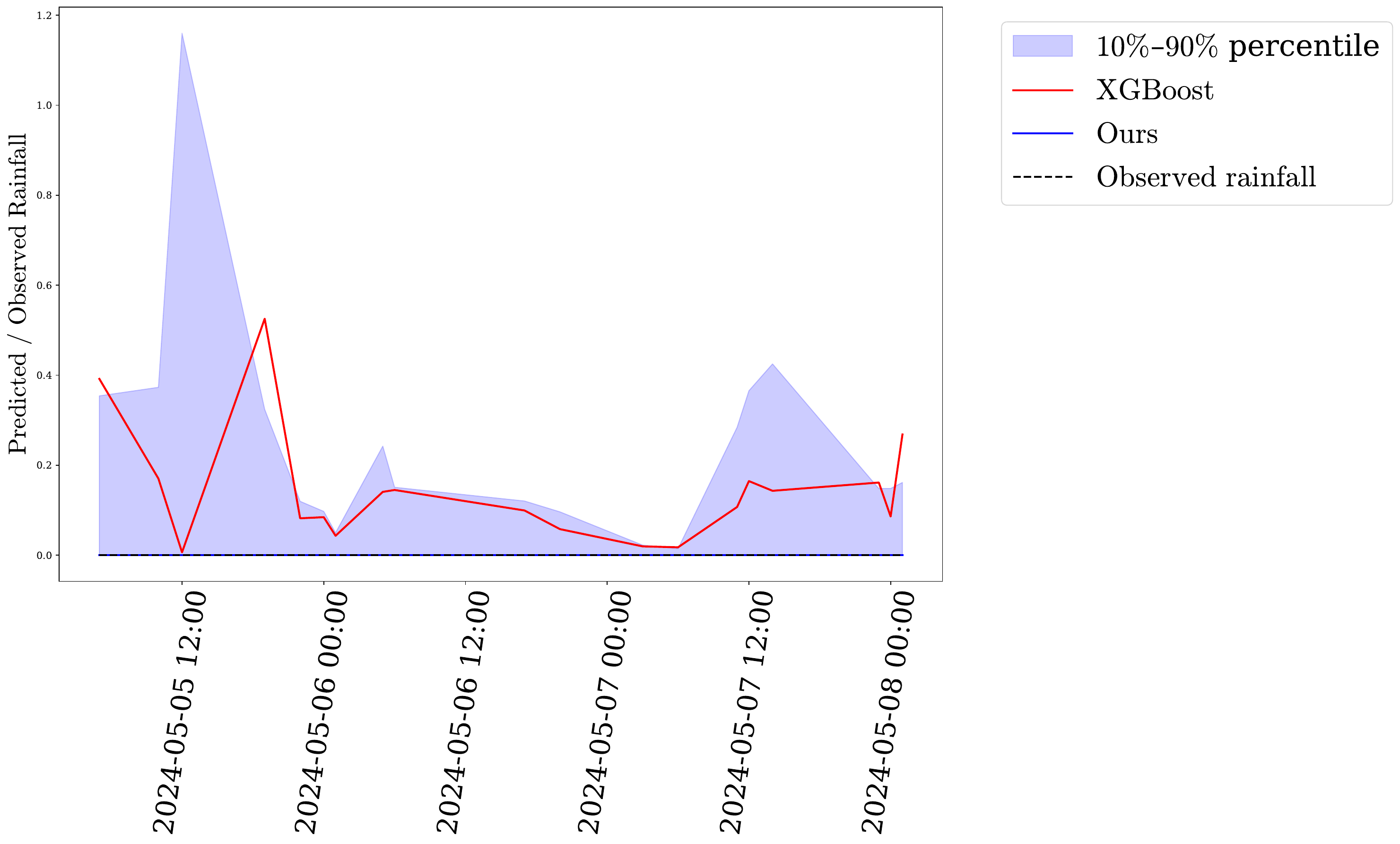}
                \subcaption{Ban Nhung Basin at session 04-05-2024 13:00:00}
                \label{fig:no_rain}
            \end{subfigure}
            % Second column
            \begin{subfigure}[t]{0.48\linewidth}
                \centering
                \includegraphics[width=\linewidth]{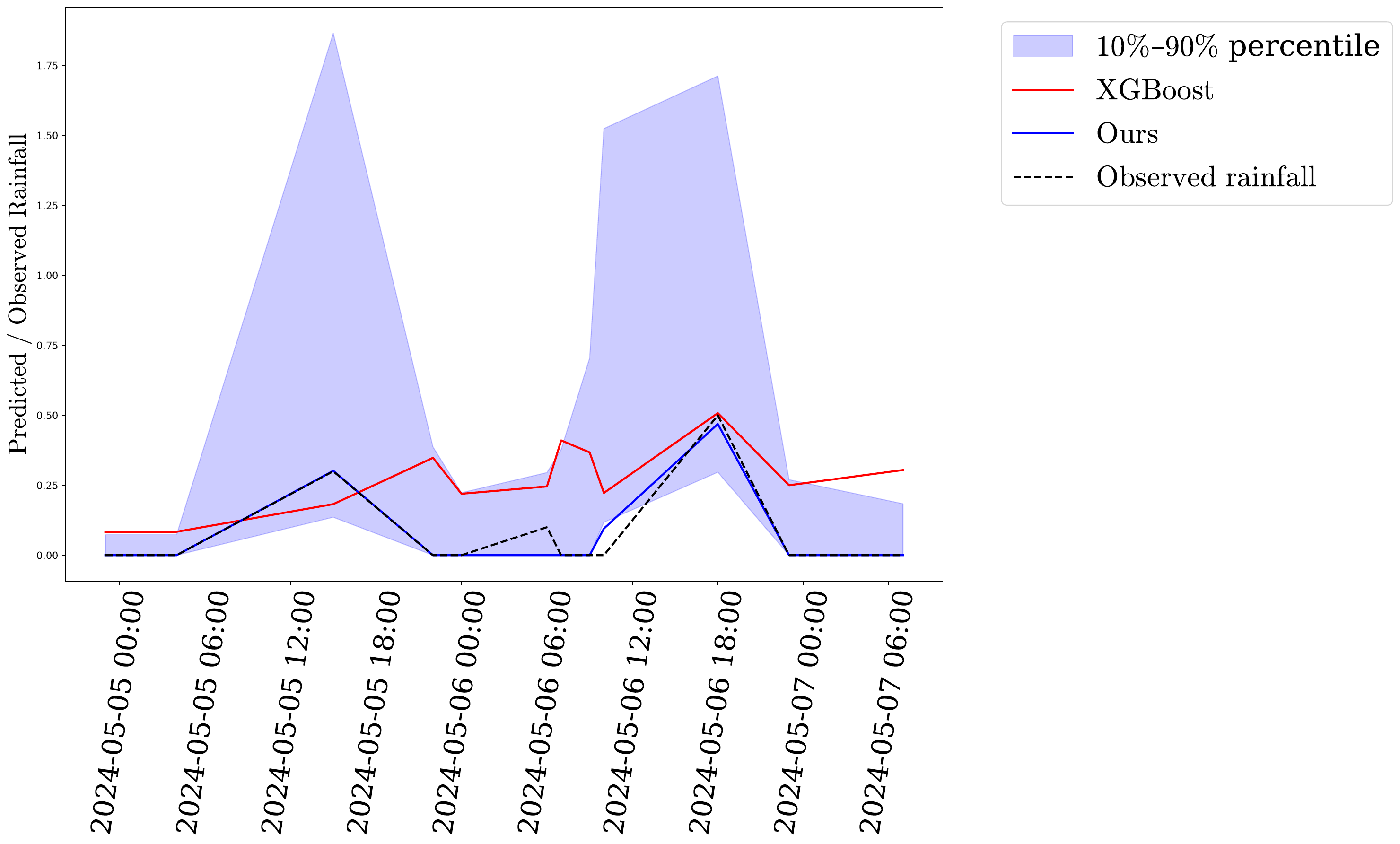}
                \subcaption{Ban Ve Basin at session 04-05-2024 19:00:00}
                \label{fig:rain}
            \end{subfigure}
            \caption{Comparison of predicted rainfall ratios between MPWE, XGBoost, and observed rainfall, with the 10--90\% percentile uncertainty interval.}
            \label{fig:rain_no_rain}
        \end{figure}
        Figure \ref{fig:rain_no_rain} illustrates the predictive capability of the models under two contrasting rainfall regimes. The Ban Nhung basin (Fig. \ref{fig:no_rain}) represents a non-rainfall scenario, where the proposed MPWE method maintains stable predictions close to zero, avoiding false alarms that are more evident in the XGBoost baseline. In contrast, the Ban Ve basin (Fig. \ref{fig:rain}) depicts an active rainfall event with multiple peaks. Here, MPWE follows the observed rainfall's timing more accurately and confines its predictions within a tighter uncertainty band, whereas XGBoost exhibits noticeable deviations. These examples highlight that MPWE can robustly discriminate between dry and wet conditions, ensuring reliable forecasts across heterogeneous hydrological regimes.

    \subsection{Hyperparameter Tuning}
        Some experiments were conducted on the Ban Nhung Basin dataset, where the proposed regime-switching combiner was tuned using a grid search implemented in \texttt{scikit-learn}. The search explored different numbers of clusters, matrix profile window sizes, and regularization strengths to identify the hyperparameters that minimize prediction error.

        \begin{figure}
            \centering
            \includegraphics[width=0.7\linewidth]{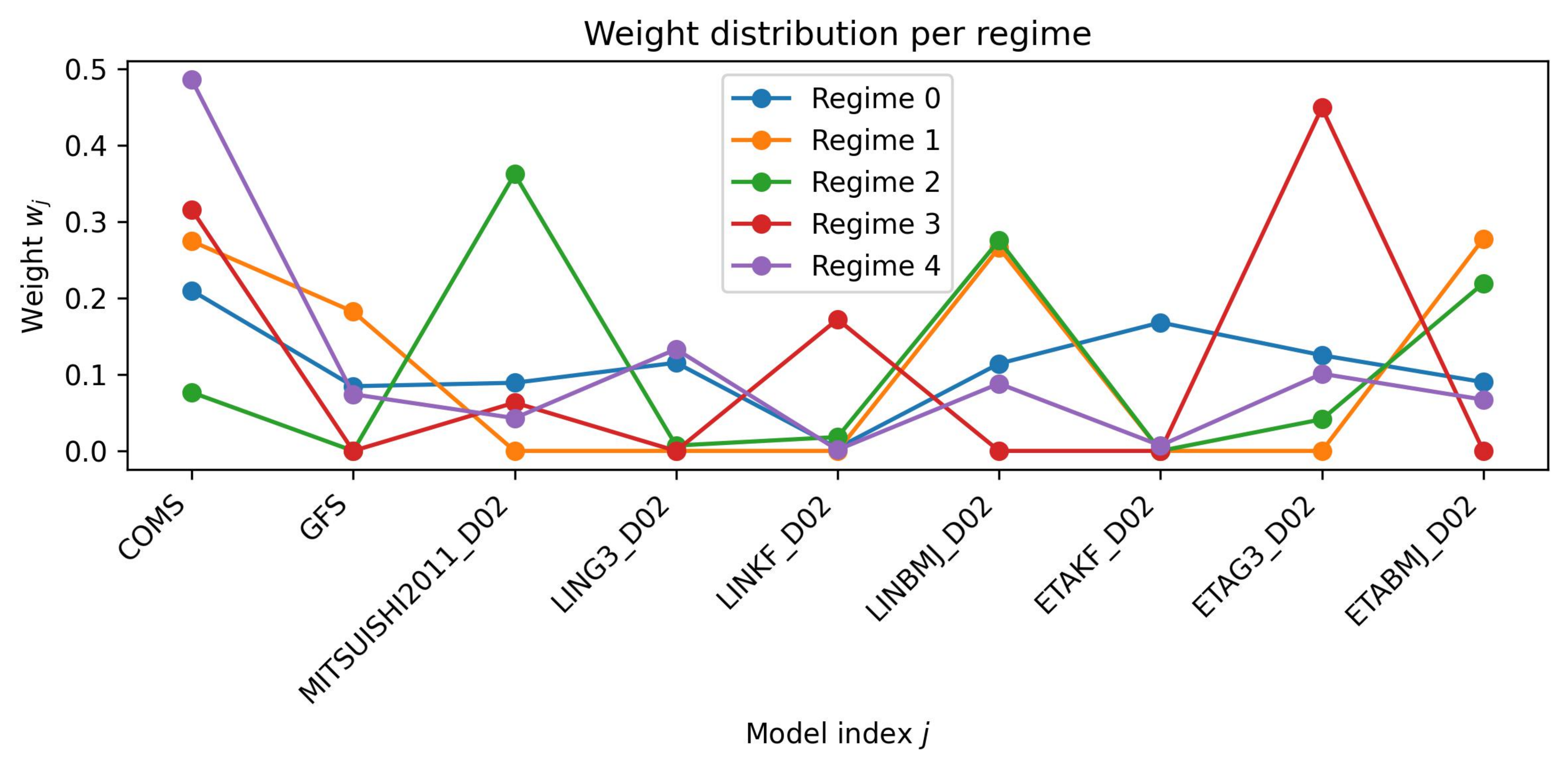}
            \caption{Regime-specific weight distributions of base models learned by the proposed combiner on the Ban Nhung Basin dataset. Each line represents one regime, showing how the contribution of individual models changes across different hydrological patterns.}
            \label{fig:weights_per_regime}
        \end{figure}
        Figure \ref{fig:weights_per_regime} shows the resulting regime-specific weight distributions under the optimal configuration. The figure highlights how the model dynamically assigns importance to different base models across regimes. For example, COMS plays a dominant role in Regime 0, while ETABMJ\_D02 contributes most strongly in Regime 3. This variability demonstrates that the model adapts to different hydrological patterns within the Ban Nhung Basin, rather than relying on a static ensemble.
\begin{figure}[htbp]
    \centering
    \begin{minipage}{0.48\linewidth}
        \centering
        \includegraphics[width=\linewidth]{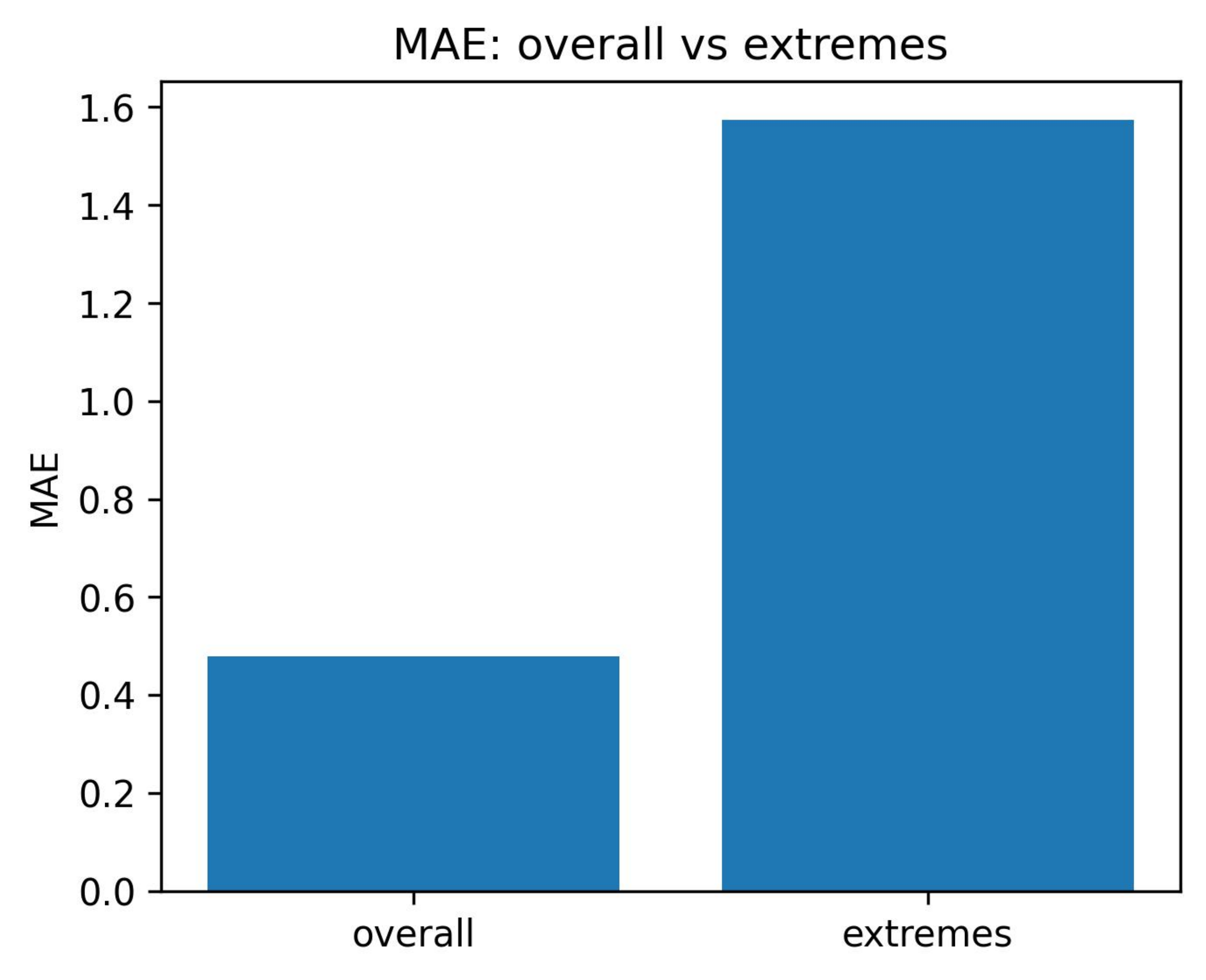}
        \caption{Comparison of mean absolute error (MAE) between overall conditions and extreme rainfall events on the Ban Nhung Basin dataset. The higher MAE in extremes reflects the inherent difficulty of predicting rare but intense events.}
        \label{fig:mae_overall_vs_extremes}
    \end{minipage}\hfill
    \begin{minipage}{0.48\linewidth}
        \centering
        \includegraphics[width=\linewidth]{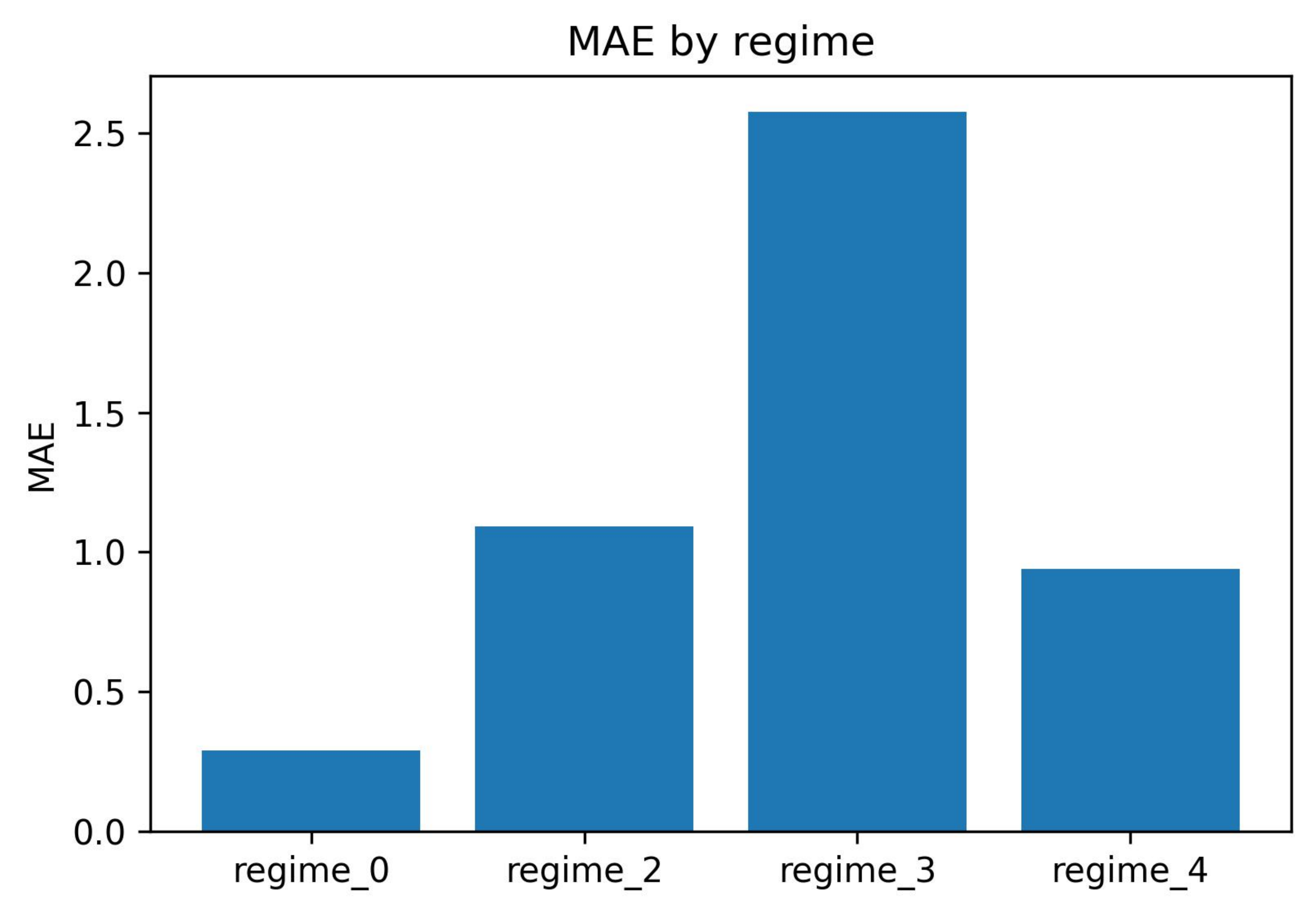}
        \caption{Mean absolute error (MAE) across regimes on the Ban Nhung Basin dataset. The proposed model achieves low error in stable regimes (e.g., Regime 0), while more volatile regimes (e.g., Regime 3) remain challenging but still benefit from the tuned regime-switching framework.}
        \label{fig:mae_by_regime}
    \end{minipage}
\end{figure}

        % \begin{figure}
        %     \centering
        %     \includegraphics[width=0.5\linewidth]{figures/mae_overall_vs_extremes.pdf}
        %     \caption{Comparison of mean absolute error (MAE) between overall conditions and extreme rainfall events on the Ban Nhung Basin dataset. The higher MAE in extremes reflects the inherent difficulty of predicting rare but intense events.}
        %     \label{fig:mae_overall_vs_extremes}
        % \end{figure}
        Figure \ref{fig:mae_overall_vs_extremes} compares the mean absolute error (MAE) between overall conditions and extreme rainfall events. As expected, extremes remain more difficult to predict, but the tuned model still achieves the lowest possible error across both settings, confirming that grid search is essential for balancing performance.
        
        % \begin{figure}
        %      \centering
        %      \includegraphics[width=0.5\linewidth]{figures/mae_by_regime.pdf}
        %      \caption{Mean absolute error (MAE) across regimes on the Ban Nhung Basin dataset. The proposed model achieves low error in stable regimes (e.g., Regime 0), while more volatile regimes (e.g., Regime 3) remain challenging but still benefit from the tuned regime-switching framework.}
        %      \label{fig:mae_by_regime}
         % \end{figure}
        Finally, Figure \ref{fig:mae_by_regime} provides the MAE distribution across individual regimes. While the model achieves very low error in stable conditions (Regime 0), it also maintains reasonable performance in more complex or volatile regimes, with Regime 3 being the most challenging. The tuned configuration nonetheless reduces errors significantly compared to untuned baselines, highlighting the effectiveness of hyperparameter optimization on the Ban Nhung Basin dataset.
\section{Discussion}
\label{sec:discussion}
    The results demonstrate that MPWE consistently outperforms both geographical models and ensemble baselines across basins and forecast horizons. By leveraging regime-aware clustering and redundancy control, MPWE adapts to temporal variability in rainfall dynamics, leading to lower mean errors and reduced variance compared to fixed-weight or regression-based ensembles. This adaptability is particularly valuable in Vietnam, where diverse climates and hydrological conditions across basins make forecasting highly challenging. While geographical models may excel in specific basins due to localized calibration, they fail to generalize, and although strong baselines such as XGBoost and QRA capture nonlinear dependencies or provide competitive short-term accuracy, they remain less robust over longer horizons.  

    Despite these advantages, several limitations remain. MPWE’s performance still depends on the quality of input forecasts, meaning that systematic biases from source models can propagate through the ensemble. Moreover, the method requires careful tuning of hyperparameters, which may limit scalability to larger datasets or real-time deployment. Finally, while MPWE improves overall stability, extreme rainfall events remain difficult to capture accurately, suggesting future work should incorporate domain knowledge, probabilistic calibration, or real-time updating to enhance resilience under severe weather conditions.  
    
\section{Conclusions}
\label{sec:conclusion}

This study demonstrates that MPWE outperforms both geographical models and ensemble baselines in rainfall forecasting across eight river basins in Vietnam. By leveraging regime-aware clustering and redundancy-aware weighting, MPWE achieves more accurate and reliable predictions over multiple horizons, showing strong adaptability to varied basin dynamics. These findings suggest that MPWE provides a robust framework for advancing rainfall forecasting in regions with complex and heterogeneous climate conditions. 

\section*{Acknowledgment}
The authors gratefully acknowledge WEATHERPLUS Solution Joint Stock Company, Vietnam, for providing the dataset that enabled the experiments in this study.

\begin{appendices}
\section{Extra Results}
\label{Appendix}
    \begin{table}[h]
        \centering
        \caption{Mean deviation of prediction errors for 2 basins: Hua Na and Khanh Khe.}
        \label{tab:error_by_basin_2_mean}
        \begin{tabular}[t]{@{}l|cccccc@{}}
            \toprule
            \textbf{Basin / Method} & \textbf{1h} & \textbf{12h} & \textbf{24h} & \textbf{48h} & \textbf{72h} & \textbf{84h} \\
            \midrule
            \multicolumn{7}{c}{\textbf{Hua Na}} \\
            \botrule

            \hline
            COMS                          & 0.50              & 2.93                & 4.67             & 7.77             & 10.76           & 12.26 \\
            GFS                           & 0.51              & 2.96                & 4.66             & 7.55             & 10.37             & 11.77 \\
            MITSUISHI2011\_D03            & 1.02              & 8.76                & 17.11            & 34.06            & 51.04            & 59.54 \\
            LINKF\_D03                    & 1.02              & 8.66                & 16.87            & 33.55            & 50.27            & 58.65 \\
            LINBMJ\_D03                   & 0.70              & 4.74                & 8.42             & 15.78             & 23.36            & 27.21 \\
            ETAKF\_D03                    & 0.82              & 6.16                & 11.62             & 22.79            & 34.13            & 39.81 \\
            ETAG3\_D03                    & 0.61              & 3.69                & 6.22             & 11.10            & 16.02            & 18.50 \\
            ETABMJ\_D03                   & 0.60              & 3.54                & 5.84             & 9.99             & 14.13             & 16.17 \\
            \hline
            Random Forest                 & 0.42              & 2.20                & 3.33             & 4.89             & 6.17             & 6.71 \\
            XGBoost                       & 0.40              & \textbf{2.05}                & \underline{3.06}             & 4.48             & \underline{5.58}             & \underline{6.07} \\
            Bayesian Model Averaging      & 0.43              & 2.37                & 3.57             & 5.45             & 6.93             & 7.56 \\
            Regression-Based Method       & 0.42              & 2.12                & 3.20             & \underline{4.70}             & 5.81             & 6.33 \\
            Simple Mean Model             & 0.43              & 2.37                & 3.57             & 5.45             & 6.93             & 7.56 \\
            Simple Median Model           & 0.42              & 2.38                & 3.66             & 5.84             & 7.66            & 8.55 \\
            Quantile Regression Averaging & \underline{0.31}              & 3.38                & 6.74             & 13.48            & 20.21            & 23.57\\ \hline
            \textbf{MPWE (ours)}          & \textbf{0.28}     & \underline{2.06}       & \textbf{3.01}    & \textbf{3.96}    & \textbf{5.05}    & \textbf{6.06} \\
            \midrule
            \multicolumn{7}{c}{\textbf{Khanh Khe}} \\
            \botrule
            COMS                          & 0.51              & 3.48                & 5.66             & 9.47             & 13.05           & 14.83 \\
            GFS                           & 0.48              & 3.06                & 4.82             & 7.52             & 9.96             & 11.13 \\
            MITSUISHI2011\_D02            & 0.76              & 6.29                & 11.83            & 23.22            & 34.82            & 40.62 \\
            LING3\_D02                    & 0.71              & 5.77                & 10.71            & 20.66            & 30.85            & 35.98 \\
            LINKF\_D02                    & 0.77              & 6.35                & 11.96            & 23.43            & 35.15            & 41.01 \\
            LINBMJ\_D02                   & 0.54              & 3.66                & 5.83             & 9.60             & 13.24            & 15.09 \\
            ETAKF\_D02                    & 0.66              & 5.12                & 9.33             & 17.81            & 26.53            & 30.91 \\
            ETAG3\_D02                    & 0.60              & 4.54                & 8.06             & 14.80            & 21.68            & 25.20 \\
            ETABMJ\_D02                   & 0.49              & 3.14                & 4.78             & 7.60             & 10.26             & 11.51 \\
            \hline
            Random Forest                 & 0.39              & 2.28                & 3.46             & 5.07             & 6.35             & 6.97 \\
            XGBoost                       & 0.37              & \underline{2.13}                & \underline{3.14}             & \underline{4.55}             & 5.49             & 5.92 \\
            Bayesian Model Averaging      & 0.37              & 2.28                & 3.34             & 4.86             & 6.07             & 6.53 \\
            Regression-Based Method       & 0.38              & 2.17                & \underline{3.14}             & 4.56             & \underline{5.48}             & \underline{5.89} \\
            Simple Mean Model             & 0.37              & 2.28                & 3.34             & 4.86             & 6.08             & 6.54 \\
            Simple Median Model           & 0.34              & 2.28                & 3.52             & 5.68             & 7.72            & 8.71 \\
            Quantile Regression Averaging & \underline{0.28}              & 2.90                & 5.74             & 11.46            & 17.19            & 20.05\\ \hline
            \textbf{MPWE (ours)}          & \textbf{0.28}     & \textbf{2.08}       & \textbf{3.13}    & \textbf{4.36}    & \textbf{5.26}    & \textbf{5.66} \\
            \botrule
            % Repeat rows
        \end{tabular}
    \end{table}

    \begin{table}[ht]
        \centering
        \caption{Standard deviation of prediction errors for 2 basins: Hua Na and Khanh Khe.}
        \label{tab:error_by_basin_2_std}
        \begin{tabular}[t]{@{}l|cccccc@{}}
            \toprule
            \textbf{Basin / Method} & \textbf{1h} & \textbf{12h} & \textbf{24h} & \textbf{48h} & \textbf{72h} & \textbf{84h} \\
            \midrule
            \multicolumn{7}{c}{\textbf{Hua Na}} \\
            \botrule

            \hline
            COMS                          & 0.87              & 2.36                & 3.44             & 5.21             & 6.88           & 7.61 \\
            GFS                           & 0.96              & 2.73                & 3.92             & 5.76             & 7.26             & 7.96 \\
            MITSUISHI2011\_D03            & 1.46              & 5.22                & 7.70             & 11.30            & 13.93            & 15.04 \\
            LINKF\_D03                    & 1.54              & 5.40                & 7.91            & 11.71            & 14.81            & 16.15 \\
            LINBMJ\_D03                   & 1.19              & 3.74                & 5.61             & 8.66             & 11.08            & 12.00 \\
            ETAKF\_D03                    & 1.21              & 4.08                & 6.19             & 9.63            & 12.20            & 13.32 \\
            ETAG3\_D03                    & 1.05              & 3.16                & 4.75             & 7.42            & 9.62            & 10.58 \\
            ETABMJ\_D03                   & 1.06              & 3.10                & 4.51             & 7.01             & 9.23             & 10.26 \\
            \hline
            Random Forest                 & \underline{0.68}              & 1.77                & \underline{2.37}             & 3.34             & 4.26             & 4.69 \\
            XGBoost                       & \underline{0.68}              & \underline{1.80}                & 2.41             & \underline{3.29}             & \textbf{4.14}             & \underline{4.45} \\
            Bayesian Model Averaging      & 0.83              & 2.22                & 3.00             & 4.19             & 5.29             & 5.81 \\
            Regression-Based Method       & 0.69              & 1.84                & 2.47             & 3.42             & 4.35             & 4.69 \\
            Simple Mean Model             & 0.83              & 2.22                & 3.00             & 4.18             & 5.28             & 5.80 \\
            Simple Median Model           & 0.83              & 2.29                & 3.17             & 4.63             & 6.11            & 6.74 \\
            Quantile Regression Averaging & 0.81              & 2.86                & 4.14             & 5.94            & 7.41            & 8.06\\ \hline
            \textbf{MPWE (ours)}          & \textbf{0.67}     & \textbf{1.23}       & \textbf{2.19}    & \textbf{2.79}    & \underline{4.36}    & \textbf{4.07} \\
            \midrule
            \multicolumn{7}{c}{\textbf{Khanh Khe}} \\
            \botrule
            COMS                          & 1.14              & 3.28                & 4.58             & 6.71             & 8.58           & 9.45 \\
            GFS                           & 1.05              & 2.93                & 4.05             & 5.84             & 7.22             & 7.80 \\
            MITSUISHI2011\_D02            & 1.36              & 4.47                & 6.86            & 10.00            & 12.13            & 13.04 \\
            LING3\_D02                    & 1.41              & 4.61                & 6.92             & 10.52            & 12.87            & 13.70 \\
            LINKF\_D02                    & 1.35              & 4.46                & 6.83            & 10.12            & 12.29            & 13.19 \\
            LINBMJ\_D02                   & 1.25              & 3.63                & 5.11             & 7.51             & 9.88            & 10.92 \\
            ETAKF\_D02                    & 1.20              & 3.81                & 5.76             & 8.78            & 10.80            & 11.61 \\
            ETAG3\_D02                    & 1.27              & 3.83                & 5.55             & 8.45            & 10.59            & 11.40 \\
            ETABMJ\_D02                   & 1.17              & 3.15                & 4.35             & 6.15             & 7.66             & 8.21 \\
            \hline
            Random Forest                 & 0.73              & \textbf{1.83}                & \underline{2.44}            & 3.52             & 4.37             & 4.66 \\
            XGBoost                       & \underline{0.72}              & 1.87                & 2.53             & \underline{3.42}             & \underline{4.18}             & \textbf{4.43} \\
            Bayesian Model Averaging      & 0.84              & 2.22                & 3.10             & 4.32             & 5.17             & 5.54 \\
            Regression-Based Method       & 0.73              & 1.89                & 2.61             & 3.54             & 4.34             & 4.60 \\
            Simple Mean Model             & 0.84              & 2.22                & 3.10             & 4.33             & 5.17             & 5.55 \\
            Simple Median Model           & 0.83              & 2.35                & 3.38             & 4.93             & 6.12            & 6.61 \\
            Quantile Regression Averaging & 0.83              & 2.87                & 4.13             & 5.91            & 7.18            & 7.70\\ \hline
            \textbf{MPWE (ours)}          & \textbf{0.70}     & \underline{2.29}       & \textbf{2.27}    & \textbf{3.77}    & \textbf{3.94}    & \underline{4.44} \\
            \botrule
            % Repeat rows
        \end{tabular}
    \end{table}
\begin{table}
        \centering
        \caption{Mean deviation of prediction errors for 2 basins: Muong Hum and Song Chay 3.}
        \label{tab:error_by_basin_3_mean}
        \begin{tabular}[t]{@{}l|cccccc@{}}
            \toprule
            \textbf{Basin / Method} & \textbf{1h} & \textbf{12h} & \textbf{24h} & \textbf{48h} & \textbf{72h} & \textbf{84h} \\
            \midrule
            \multicolumn{7}{c}{\textbf{Muong Hum}} \\
            \botrule

            \hline
            WRF84H                        & 0.70              & 5.46                & 9.16             & 15.98            & 22.78            & 26.19 \\
            COMS                          & 0.59              & 4.29                & 7.01             & 11.65            & 16.18            & 18.51 \\
            GFS                           & 0.53              & 3.65                & 5.86             & 9.10             & 12.00            & 13.42 \\
            MITSUISHI2011\_D01            & 0.98              & 8.77                & 16.29            & 31.54            & 46.89            & 54.54 \\
            MITSUISHI2021                 & 0.84              & 7.03                & 12.61            & 23.48            & 34.61            & 40.23 \\
            \hline
            Random Forest                 & 0.50              & 3.46                & 5.52             & 8.51             & 10.91            & 12.11 \\
            XGBoost                       & 0.53              & 3.72                & 5.93             & 9.07             & 11.78            & 13.11 \\
            Bayesian Model Averaging      & 0.41              & 2.98                & \underline{4.73}             & \underline{7.36}             & \textbf{9.65}             & \underline{10.76} \\
            Regression-Based Method       & 0.48              & 3.22                & 5.05             & 7.72             & 9.84             & 10.89 \\
            Simple Mean Model             & 0.41              & 2.98                & \underline{4.73}             & 7.37             & \textbf{9.65}             & \underline{10.76} \\
            Simple Median Model           & 0.38              & \underline{2.87}                & 4.76             & 8.00             & 11.11            & 12.66 \\
            Quantile Regression Averaging & \underline{0.31}              & 3.69                & 7.37             & 14.75            & 22.14            & 25.84 \\
            \hline
            \textbf{MPWE (ours)}          & \textbf{0.27}     & \textbf{2.09}       & \textbf{4.71}    & \textbf{7.07}    & \underline{10.34}   & \textbf{9.98} \\
            \midrule
            \multicolumn{7}{c}{\textbf{Song Chay 3}} \\
           \botrule
            WRF84H                        & 0.93              & 6.31                & 9.68             & 14.73            & 18.19            & 19.49 \\
            COMS                          & 0.81              & 5.27                & 7.94             & 11.87            & 14.52            & 15.50 \\
            GFS                           & 0.77              & 5.03                & 7.64             & 11.53            & 14.37            & 15.69 \\
            MITSUISHI2011\_D01            & 1.47              & 11.57               & 21.11            & 40.23            & 59.78            & 69.63 \\
            MITSUISHI2021                 & 1.45              & 11.36               & 20.71            & 39.52            & 58.57            & 68.20 \\
            \hline
            Random Forest                 & 0.85              & 5.35                & 8.09             & 12.03            & 14.65            & 15.89 \\
            XGBoost                       & 0.80              & \underline{4.93}                & \underline{7.36}             & 10.94            & \underline{13.19}            & \underline{14.14} \\
            Bayesian Model Averaging      & 0.63              & 5.00                & 8.73             & 16.08            & 23.37            & 27.03 \\
            Regression-Based Method       & 0.82              & 4.97                & 7.44             & \underline{10.93}            & 13.27            & 14.23 \\
            Simple Mean Model             & 0.63              & 5.00                & 8.73             & 16.09            & 23.38            & 27.04 \\
            Simple Median Model           & 0.60              & 5.25                & 9.60             & 18.35            & 27.21            & 31.69 \\
            Quantile Regression Averaging & \textbf{0.55}              & 6.61                & 13.21            & 26.45            & 39.69            & 46.32 \\ 
            \hline
            \textbf{MPWE (ours)}          & \textbf{0.54}     & \textbf{4.84}       & \textbf{7.25}    & \textbf{9.93}   & \textbf{11.56}   & \textbf{13.91} \\
            \botrule
            % Repeat rows
        \end{tabular}
    \end{table}
\begin{table}
        \centering
        \caption{Standard deviation of prediction errors for 2 basins: Muong Hum and Song Chay 3.}
        \label{tab:error_by_basin_3_std}
        \begin{tabular}[t]{@{}l|cccccc@{}}
            \toprule
            \textbf{Basin / Method} & \textbf{1h} & \textbf{12h} & \textbf{24h} & \textbf{48h} & \textbf{72h} & \textbf{84h} \\
            \midrule
            \multicolumn{7}{c}{\textbf{Muong Hum}} \\
            \botrule
            \hline
            WRF84H                        & 2.72              & 8.77                & 12.17            & 16.87            & 20.55            & 22.12 \\
            COMS                          & 2.45              & 8.00                & 11.09            & 15.28            & 18.43            & 19.72 \\
            GFS                           & 2.39              & 7.88                & 10.91            & 14.95            & 17.95            &\textbf{19.25} \\
            MITSUISHI2011\_D01            & 2.99              & 9.76                & 13.86            & 19.20            & 22.68            & 24.13 \\
            MITSUISHI2021                 & 2.86              & 9.23                & 13.00            & 17.84            & 21.01            & 22.31 \\
            \hline
            Random Forest                 & 2.54              & 8.34                & 11.45            & 15.57            & 18.73            & 20.10 \\
            XGBoost                       & 2.53              & 8.21                & 11.41            & 15.72            & 18.72            & 20.03 \\
            Bayesian Model Averaging      & 2.40              & 8.09                & 11.34            & 15.75            & 19.15            & 20.67 \\
            Regression-Based Method       & \textbf{2.35}              & \underline{7.81}                & \textbf{10.89}            & \underline{15.01}            & \underline{18.13}            & 19.47 \\
            Simple Mean Model             & 2.40              & 8.09                & 11.35            & 15.75            & 19.15            & 20.68 \\
            Simple Median Model           & 2.39              & 8.17                & 11.53            & 16.15            & 19.76            & 21.39 \\
            Quantile Regression Averaging & 2.41              & 8.47                & 12.04            & 16.92            & 20.64            & 22.26 \\
            \textbf{MPWE (ours)}          & \underline{2.38}     & \textbf{7.15}       & \underline{11.49}   & \textbf{15.13}   & \textbf{17.76}   & \underline{19.90} \\
            \midrule
            \multicolumn{7}{c}{\textbf{Song Chay 3}} \\
            \botrule
            WRF84H                        & 2.38              & 6.37                & 8.57             & 11.67            & 14.07            & 15.17 \\
            COMS                          & 1.98              & 5.38                & 7.31             & 9.88             & 11.70            & 12.52 \\
            GFS                           & 1.95              & 5.58                & 7.68             & 10.52            & 12.80            & 13.72 \\
            MITSUISHI2011\_D01            & 2.76              & 8.67                & 12.88            & 19.82            & 24.60            & 26.63 \\
            MITSUISHI2021                 & 2.92              & 9.08                & 13.66            & 21.00            & 26.34            & 28.27 \\
            \hline
            Random Forest                 & 1.88              & 5.02                & 6.69             & 9.16             & 11.27            & 12.04 \\
            XGBoost                       & \underline{1.81}              & \underline{4.93}                & \underline{6.68}             & \underline{8.99}             & \underline{10.86}            & \underline{11.52} \\
            Bayesian Model Averaging      & 1.93              & 6.29                & 9.20             & 13.45            & 16.67            & 18.01 \\
            Regression-Based Method       & 1.81              & 4.94                & 6.76             & 9.19             & 11.03            & 11.73 \\
            Simple Mean Model             & 1.93              & 6.29                & 9.21             & 13.45            & 16.67            & 18.01 \\
            Simple Median Model           & 1.95              & 6.55                & 9.64             & 14.10            & 17.34            & 18.62 \\
            Quantile Regression Averaging & 2.04              & 7.20                & 10.36            & 14.67            & 17.83            & 19.11 \\
            \hline
            \textbf{MPWE (ours)}          & \textbf{1.71}     & \textbf{4.74}       & \textbf{6.00}    & \textbf{8.17}   & \textbf{10.33}   & \textbf{10.60} \\
            \botrule
            % Repeat rows
        \end{tabular}
    \end{table}

\begin{table}
        \centering
        \caption{Mean deviation of prediction errors for 2 basins: Song Chung and Thac Xang.}
        \label{tab:error_by_basin_4_mean}
        \begin{tabular}[t]{@{}l|cccccc@{}}
            \toprule
            \textbf{Basin / Method} & \textbf{1h} & \textbf{12h} & \textbf{24h} & \textbf{48h} & \textbf{72h} & \textbf{84h} \\
            \midrule
            \multicolumn{7}{c}{\textbf{Song Chung}} \\
            \botrule

            \hline
            WRF84H                        & 0.75              & 5.46                & 8.43             & 12.90            & 16.73            & 18.63 \\
            COMS                          & 0.73              & 4.92                & 7.33             & 11.09            & 14.54            & 16.13 \\
            GFS                           & 0.77              & 5.07                & 7.52             & 11.16            & 14.35            & 15.71 \\
            MITSUISHI2011\_D01            & 1.91              & 17.20               & 33.31            & 66.23            & 99.38            & 115.94 \\
            MITSUISHI2021                 & 1.59              & 13.09               & 24.63            & 48.52            & 72.76            & 84.90 \\
            \hline
            Random Forest                 & 0.85              & 5.40                & 7.93             & 11.62            & 14.85            & 16.30 \\
            XGBoost                       & 0.81              & 5.07                & 7.43             & 10.90            & 13.79            & 15.03 \\
            Bayesian Model Averaging      & 0.62              & \underline{4.81}                & 8.24             & 14.72            & 21.26            & 24.55 \\
            Regression-Based Method       & 0.82              & 5.05                & \underline{7.35}             & \underline{10.65}            & \underline{13.49}            & \underline{14.73} \\
            Simple Mean Model             & 0.62              & \underline{4.81}                & 8.25             & 14.74            & 21.30            & 24.60 \\
            Simple Median Model           & 0.60              & 4.95                & 8.84             & 16.45            & 24.22            & 28.16 \\
            Quantile Regression Averaging & \underline{0.54}              & 6.37                & 12.74            & 25.48            & 38.20            & 44.57 \\
            \hline
            \textbf{MPWE (ours)}          & \textbf{0.45}     & \textbf{4.68}       & \textbf{7.25}    & \textbf{10.06}   & \textbf{13.47}   & \textbf{14.15} \\
            \midrule
            \multicolumn{7}{c}{\textbf{Thac Xang}} \\
            \botrule
            COMS                          & 0.77              & 5.08                & 7.97             & 12.01            & 15.13            & 16.51 \\
            GFS                           & 0.74              & \textbf{4.72}                & 7.34             & 11.02            & 13.74            & 15.01 \\
            MITSUISHI2011\_D02            & 1.28              & 9.16                & 15.46            & 27.56            & 39.84            & 45.95 \\
            LING3\_D02                    & 1.11              & 7.56                & 12.29            & 20.59            & 28.87            & 32.88 \\
            LINKF\_D02                    & 1.30              & 9.29                & 15.77            & 28.30            & 40.91            & 47.21 \\
            LINBMJ\_D02                   & 0.84              & 5.45                & 8.27             & 11.88            & 14.66            & 15.83 \\
            ETAKF\_D02                    & 1.12              & 7.49                & 12.11            & 20.08            & 27.96            & 31.86 \\
            ETAG3\_D02                    & 0.97              & 6.26                & 9.69             & 15.04            & 19.86            & 22.15 \\
            ETABMJ\_D02                   & 0.78              & 5.03                & 7.69             & 11.39            & 14.57            & 16.02 \\
            \hline
            Random Forest                 & 0.78              & 4.84                & 7.45             & 11.06            & 14.19            & 15.61 \\
            XGBoost                       & 0.77              & 4.76                & \underline{7.22}             & \underline{10.36}            & \underline{12.94}            & \underline{14.10} \\
            Bayesian Model Averaging      & 0.62              & 4.86                & 8.62             & 15.90            & 23.39            & 27.18 \\
            Regression-Based Method       & 0.81              & 4.89                & 7.55             & 10.97            & 13.41            & 14.49 \\
            Simple Mean Model             & 0.63              & 4.85                & 8.59             & 15.81            & 23.23            & 26.98 \\
            Simple Median Model           & 0.61              & 5.11                & 9.40             & 17.91            & 26.67            & 31.07 \\
            Quantile Regression Averaging & \textbf{0.55}              & 5.98                & 11.92            & 23.86            & 35.84            & 41.81\\ \hline
            \textbf{MPWE (ours)}          & \underline{0.59}     & \textbf{4.72}       & \textbf{7.01}    & \textbf{10.24}   & \textbf{12.65}   & \textbf{13.88} \\
            \botrule
            % Repeat rows
        \end{tabular}
    \end{table}

\begin{table}
        \centering
        \caption{Standard deviation of prediction errors for 2 basins: Song Chung and Thac Xang.}
        \label{tab:error_by_basin_4_std}
        \begin{tabular}[t]{@{}l|cccccc@{}}
            \toprule
            \textbf{Basin / Method} & \textbf{1h} & \textbf{12h} & \textbf{24h} & \textbf{48h} & \textbf{72h} & \textbf{84h} \\
            \midrule
            \multicolumn{7}{c}{\textbf{Song Chung}} \\
            \botrule

            \hline
            WRF84H                        & 2.22              & 6.06                & 8.00             & 11.07            & 13.82            & 15.09 \\
            COMS                          & 1.96              & 5.34                & 7.16             & 10.00            & 12.42            & 13.61 \\
            GFS                           & 1.95              & 5.14                & 6.79             & 9.52             & 11.81            & 12.95 \\
            MITSUISHI2011\_D01            & 2.95              & 10.29               & 15.63            & 22.88            & 28.28            & 30.75 \\
            MITSUISHI2021                 & 2.65              & 8.68                & 13.06            & 19.38            & 23.62            & 25.35 \\
            \hline
            Random Forest                 & 1.88              & 4.66                & 6.08             & 8.52             & 10.73            & 11.78 \\
            XGBoost                       & 1.80              & 4.57                & \underline{5.96}             & \underline{8.25}             & 10.31            & 11.29 \\
            Bayesian Model Averaging      & 1.91              & 5.95                & 8.35             & 12.42            & 16.04            & 17.72 \\
            Regression-Based Method       & \underline{1.79}              & \underline{4.53}                & \underline{5.96}             & 8.32             & \underline{10.29}            & \underline{11.24} \\
            Simple Mean Model             & 1.91              & 5.95                & 8.36             & 12.43            & 16.06            & 17.74 \\
            Simple Median Model           & 1.92              & 6.16                & 8.73             & 12.99            & 16.64            & 18.25 \\
            Quantile Regression Averaging & 2.02              & 6.99                & 9.80             & 14.01            & 17.56            & 19.15 \\
            \hline
            \textbf{MPWE (ours)}          & \textbf{1.77}     & \textbf{4.26}       & \textbf{5.90}    & \textbf{7.70}   & \textbf{10.13}   & \textbf{10.73} \\
            \midrule
            \multicolumn{7}{c}{\textbf{Thac Xang}} \\
           \botrule
            COMS                          & 2.08              & 5.65                & 7.34             & 9.88             & 11.93            & 12.81 \\
            GFS                           & 1.99              & 5.57                & 7.36             & 10.02            & 12.17            & 13.08 \\
            MITSUISHI2011\_D02            & 2.32              & 6.01                & 9.16             & 14.49            & 19.20            & 21.38 \\
            LING3\_D02                    & 2.30              & 5.70                & 8.03             & 12.53            & 16.54            & 18.53 \\
            LINKF\_D02                    & 2.37              & 6.20                & 9.43             & 14.76            & 19.70            & 21.91 \\
            LINBMJ\_D02                   & 2.16              & 5.82                & 7.51             & 9.92             & 11.67            & 12.36 \\
            ETAKF\_D02                    & 2.18              & 5.28                & 7.52             & 11.90            & 15.84            & 17.67 \\
            ETAG3\_D02                    & 2.20              & 5.45                & 7.16             & 10.42            & 13.62            & 15.02 \\
            ETABMJ\_D02                   & 2.11              & 5.87                & 7.75             & 10.73            & 13.21            & 14.25 \\
            \hline
            Random Forest                 & 1.73              & 4.56                & \underline{5.96}             & 8.28             & 9.71             & 10.22 \\
            XGBoost                       & \underline{1.72}              & \underline{4.53}                & \underline{5.96}             & \underline{8.17}             & \underline{9.43}             & \underline{9.85} \\
            Bayesian Model Averaging      & 2.01              & 6.38                & 9.13             & 13.23            & 16.42            & 17.72 \\
            Regression-Based Method       & 1.86              & 4.87                & 6.20             & \underline{8.17}             & 9.68             & 10.23 \\
            Simple Mean Model             & 2.01              & 6.37                & 9.11             & 13.21            & 16.42            & 17.73 \\
            Simple Median Model           & 2.02              & 6.56                & 9.47             & 13.76            & 16.92            & 18.21 \\
            Quantile Regression Averaging & 2.02              & 6.89                & 9.80             & 13.70            & 16.49            & 17.63\\ \hline
            \textbf{MPWE (ours)}          & \textbf{1.21}     & \textbf{4.49}       & \textbf{5.34}    & \textbf{8.44}   & \textbf{8.54}   & \textbf{9.79} \\
            \botrule
            % Repeat rows
        \end{tabular}
    \end{table}



\end{appendices}

\clearpage
\bibliography{references}% common bib file


\begin{thebibliography}{27}
\ifx \bisbn   \undefined \def \bisbn  #1{ISBN #1}\fi
\ifx \binits  \undefined \def \binits#1{#1}\fi
\ifx \bauthor  \undefined \def \bauthor#1{#1}\fi
\ifx \batitle  \undefined \def \batitle#1{#1}\fi
\ifx \bjtitle  \undefined \def \bjtitle#1{#1}\fi
\ifx \bvolume  \undefined \def \bvolume#1{\textbf{#1}}\fi
\ifx \byear  \undefined \def \byear#1{#1}\fi
\ifx \bissue  \undefined \def \bissue#1{#1}\fi
\ifx \bfpage  \undefined \def \bfpage#1{#1}\fi
\ifx \blpage  \undefined \def \blpage #1{#1}\fi
\ifx \burl  \undefined \def \burl#1{\textsf{#1}}\fi
\ifx \doiurl  \undefined \def \doiurl#1{\url{https://doi.org/#1}}\fi
\ifx \betal  \undefined \def \betal{\textit{et al.}}\fi
\ifx \binstitute  \undefined \def \binstitute#1{#1}\fi
\ifx \binstitutionaled  \undefined \def \binstitutionaled#1{#1}\fi
\ifx \bctitle  \undefined \def \bctitle#1{#1}\fi
\ifx \beditor  \undefined \def \beditor#1{#1}\fi
\ifx \bpublisher  \undefined \def \bpublisher#1{#1}\fi
\ifx \bbtitle  \undefined \def \bbtitle#1{#1}\fi
\ifx \bedition  \undefined \def \bedition#1{#1}\fi
\ifx \bseriesno  \undefined \def \bseriesno#1{#1}\fi
\ifx \blocation  \undefined \def \blocation#1{#1}\fi
\ifx \bsertitle  \undefined \def \bsertitle#1{#1}\fi
\ifx \bsnm \undefined \def \bsnm#1{#1}\fi
\ifx \bsuffix \undefined \def \bsuffix#1{#1}\fi
\ifx \bparticle \undefined \def \bparticle#1{#1}\fi
\ifx \barticle \undefined \def \barticle#1{#1}\fi
\bibcommenthead
\ifx \bconfdate \undefined \def \bconfdate #1{#1}\fi
\ifx \botherref \undefined \def \botherref #1{#1}\fi
\ifx \url \undefined \def \url#1{\textsf{#1}}\fi
\ifx \bchapter \undefined \def \bchapter#1{#1}\fi
\ifx \bbook \undefined \def \bbook#1{#1}\fi
\ifx \bcomment \undefined \def \bcomment#1{#1}\fi
\ifx \oauthor \undefined \def \oauthor#1{#1}\fi
\ifx \citeauthoryear \undefined \def \citeauthoryear#1{#1}\fi
\ifx \endbibitem  \undefined \def \endbibitem {}\fi
\ifx \bconflocation  \undefined \def \bconflocation#1{#1}\fi
\ifx \arxivurl  \undefined \def \arxivurl#1{\textsf{#1}}\fi
\csname PreBibitemsHook\endcsname

%%% 1
\bibitem[\protect\citeauthoryear{Arias et~al.}{2021}]{RN4}
\begin{bbook}
\bauthor{\bsnm{Arias}, \binits{P.A.}},
\bauthor{\bsnm{Bellouin}, \binits{N.}},
\bauthor{\bsnm{Coppola}, \binits{E.}},
\bauthor{\bsnm{Jones}, \binits{R.G.}},
\bauthor{\bsnm{Krinner}, \binits{G.}},
\bauthor{\bsnm{Marotzke}, \binits{J.}},
\bauthor{\bsnm{Naik}, \binits{V.}},
\bauthor{\bsnm{Palmer}, \binits{M.D.}},
\bauthor{\bsnm{Plattner}, \binits{G.-K.}},
\bauthor{\bsnm{Rogelj}, \binits{J.}},
\bauthor{\bsnm{Rojas}, \binits{M.}},
\bauthor{\bsnm{Sillmann}, \binits{J.}},
\bauthor{\bsnm{Storelvmo}, \binits{T.}},
\bauthor{\bsnm{Thorne}, \binits{P.W.}},
\bauthor{\bsnm{Trewin}, \binits{B.}},
\bauthor{\bsnm{Achuta~Rao}, \binits{K.}},
\bauthor{\bsnm{Adhikary}, \binits{B.}},
\bauthor{\bsnm{Allan}, \binits{R.P.}},
\bauthor{\bsnm{Armour}, \binits{K.}},
\bauthor{\bsnm{Bala}, \binits{G.}},
\bauthor{\bsnm{Barimalala}, \binits{R.}},
\bauthor{\bsnm{Berger}, \binits{S.}},
\bauthor{\bsnm{Canadell}, \binits{J.G.}},
\bauthor{\bsnm{Cassou}, \binits{C.}},
\bauthor{\bsnm{Cherchi}, \binits{A.}},
\bauthor{\bsnm{Collins}, \binits{W.}},
\bauthor{\bsnm{Collins}, \binits{W.D.}},
\bauthor{\bsnm{Connors}, \binits{S.L.}},
\bauthor{\bsnm{Corti}, \binits{S.}},
\bauthor{\bsnm{Cruz}, \binits{F.}},
\bauthor{\bsnm{Dentener}, \binits{F.J.}},
\bauthor{\bsnm{Dereczynski}, \binits{C.}},
\bauthor{\bsnm{Di~Luca}, \binits{A.}},
\bauthor{\bsnm{Diongue~Niang}, \binits{A.}},
\bauthor{\bsnm{Doblas-Reyes}, \binits{F.J.}},
\bauthor{\bsnm{Dosio}, \binits{A.}},
\bauthor{\bsnm{Douville}, \binits{H.}},
\bauthor{\bsnm{Engelbrecht}, \binits{F.}},
\bauthor{\bsnm{Eyring}, \binits{V.}},
\bauthor{\bsnm{Fischer}, \binits{E.}},
\bauthor{\bsnm{Forster}, \binits{P.}},
\bauthor{\bsnm{Fox-Kemper}, \binits{B.}},
\bauthor{\bsnm{Fuglestvedt}, \binits{J.S.}},
\bauthor{\bsnm{Fyfe}, \binits{J.C.}},
\bauthor{\bsnm{Gillett}, \binits{N.P.}},
\bauthor{\bsnm{Goldfarb}, \binits{L.}},
\bauthor{\bsnm{Gorodetskaya}, \binits{I.}},
\bauthor{\bsnm{Gutierrez}, \binits{J.M.}},
\bauthor{\bsnm{Hamdi}, \binits{R.}},
\bauthor{\bsnm{Hawkins}, \binits{E.}},
\bauthor{\bsnm{Hewitt}, \binits{H.T.}},
\bauthor{\bsnm{Hope}, \binits{P.}},
\bauthor{\bsnm{Islam}, \binits{A.S.}},
\bauthor{\bsnm{Jones}, \binits{C.}},
\bauthor{\bsnm{Kaufman}, \binits{D.S.}},
\bauthor{\bsnm{Kopp}, \binits{R.E.}},
\bauthor{\bsnm{Kosaka}, \binits{Y.}},
\bauthor{\bsnm{Kossin}, \binits{J.}},
\bauthor{\bsnm{Krakovska}, \binits{S.}},
\bauthor{\bsnm{Lee}, \binits{J.-Y.}},
\bauthor{\bsnm{Li}, \binits{J.}},
\bauthor{\bsnm{Mauritsen}, \binits{T.}},
\bauthor{\bsnm{Maycock}, \binits{T.K.}},
\bauthor{\bsnm{Meinshausen}, \binits{M.}},
\bauthor{\bsnm{Min}, \binits{S.-K.}},
\bauthor{\bsnm{Monteiro}, \binits{P.M.S.}},
\bauthor{\bsnm{Ngo-Duc}, \binits{T.}},
\bauthor{\bsnm{Otto}, \binits{F.}},
\bauthor{\bsnm{Pinto}, \binits{I.}},
\bauthor{\bsnm{Pirani}, \binits{A.}},
\bauthor{\bsnm{Raghavan}, \binits{K.}},
\bauthor{\bsnm{Ranasinghe}, \binits{R.}},
\bauthor{\bsnm{Ruane}, \binits{A.C.}},
\bauthor{\bsnm{Ruiz}, \binits{L.}},
\bauthor{\bsnm{Sallée}, \binits{J.-B.}},
\bauthor{\bsnm{Samset}, \binits{B.H.}},
\bauthor{\bsnm{Sathyendranath}, \binits{S.}},
\bauthor{\bsnm{Seneviratne}, \binits{S.I.}},
\bauthor{\bsnm{Sörensson}, \binits{A.A.}},
\bauthor{\bsnm{Szopa}, \binits{S.}},
\bauthor{\bsnm{Takayabu}, \binits{I.}},
\bauthor{\bsnm{Tréguier}, \binits{A.-M.}},
\bauthor{\bsnm{Hurk}, \binits{B.}},
\bauthor{\bsnm{Vautard}, \binits{R.}},
\bauthor{\bsnm{Schuckmann}, \binits{K.}},
\bauthor{\bsnm{Zaehle}, \binits{S.}},
\bauthor{\bsnm{Zhang}, \binits{X.}},
\bauthor{\bsnm{Zickfeld}, \binits{K.}}:
In: \beditor{\bsnm{Masson-Delmotte}, \binits{V.}},
\beditor{\bsnm{Zhai}, \binits{P.}},
\beditor{\bsnm{Pirani}, \binits{A.}},
\beditor{\bsnm{Connors}, \binits{S.L.}},
\beditor{\bsnm{Péan}, \binits{C.}},
\beditor{\bsnm{Berger}, \binits{S.}},
\beditor{\bsnm{Caud}, \binits{N.}},
\beditor{\bsnm{Chen}, \binits{Y.}},
\beditor{\bsnm{Goldfarb}, \binits{L.}},
\beditor{\bsnm{Gomis}, \binits{M.I.}},
\beditor{\bsnm{Huang}, \binits{M.}},
\beditor{\bsnm{Leitzell}, \binits{K.}},
\beditor{\bsnm{Lonnoy}, \binits{E.}},
\beditor{\bsnm{Matthews}, \binits{J.B.R.}},
\beditor{\bsnm{Maycock}, \binits{T.K.}},
\beditor{\bsnm{Waterfield}, \binits{T.}},
\beditor{\bsnm{Yelekçi}, \binits{O.}},
\beditor{\bsnm{Yu}, \binits{R.}},
\beditor{\bsnm{Zhou}, \binits{B.}} (eds.)
\bbtitle{Technical Summary},
pp. \bfpage{33}--\blpage{144}.
\bpublisher{Cambridge University Press},
\blocation{Cambridge, United Kingdom and New York, NY, USA}
(\byear{2021}).
\doiurl{10.1017/9781009157896.002}
\end{bbook}
\endbibitem

%%% 2
\bibitem[\protect\citeauthoryear{Allan and Soden}{2008}]{1160787}
\begin{barticle}
\bauthor{\bsnm{Allan}, \binits{R.P.}},
\bauthor{\bsnm{Soden}, \binits{B.J.}}:
\batitle{Atmospheric warming and the amplification of precipitation extremes}.
\bjtitle{Science}
\bvolume{321}(\bissue{5895}),
\bfpage{1481}--\blpage{1484}
(\byear{2008})
\doiurl{10.1126/science.1160787}
{\href{https://arxiv.org/abs/https://www.science.org/doi/pdf/10.1126/science.1160787}{{https://www.science.org/doi/pdf/10.1126/science.1160787}}}
\end{barticle}
\endbibitem

%%% 3
\bibitem[\protect\citeauthoryear{Giorgi and Mearns}{2002}]{Giorgi2002-tt}
\begin{barticle}
\bauthor{\bsnm{Giorgi}, \binits{F.}},
\bauthor{\bsnm{Mearns}, \binits{L.O.}}:
\batitle{Calculation of average, uncertainty range, and reliability of regional climate changes from {AOGCM} simulations via the ``reliability ensemble averaging'' ({REA}) method}.
\bjtitle{J. Clim.}
\bvolume{15}(\bissue{10}),
\bfpage{1141}--\blpage{1158}
(\byear{2002})
\end{barticle}
\endbibitem

%%% 4
\bibitem[\protect\citeauthoryear{Chen et~al.}{2011}]{chen2011uncertainty}
\begin{barticle}
\bauthor{\bsnm{Chen}, \binits{J.}},
\bauthor{\bsnm{Brissette}, \binits{F.P.}},
\bauthor{\bsnm{Leconte}, \binits{R.}}:
\batitle{Uncertainty of downscaling method in quantifying the impact of climate change on hydrology}.
\bjtitle{Journal of hydrology}
\bvolume{401}(\bissue{3-4}),
\bfpage{190}--\blpage{202}
(\byear{2011})
\end{barticle}
\endbibitem

%%% 5
\bibitem[\protect\citeauthoryear{Maraun}{2016}]{maraun2016bias}
\begin{barticle}
\bauthor{\bsnm{Maraun}, \binits{D.}}:
\batitle{Bias correcting climate change simulations-a critical review}.
\bjtitle{Current Climate Change Reports}
\bvolume{2}(\bissue{4}),
\bfpage{211}--\blpage{220}
(\byear{2016})
\end{barticle}
\endbibitem

%%% 6
\bibitem[\protect\citeauthoryear{Tuyet et~al.}{2019}]{tuyet2019performance}
\begin{barticle}
\bauthor{\bsnm{Tuyet}, \binits{N.T.}},
\bauthor{\bsnm{Thanh}, \binits{N.D.}},
\bauthor{\bsnm{Van~Tan}, \binits{P.}}:
\batitle{Performance of seaclid/cordex-sea multi-model experiments in simulating temperature and rainfall in vietnam}.
\bjtitle{Vietnam Journal of Earth Sciences}
\bvolume{41}(\bissue{4}),
\bfpage{374}--\blpage{387}
(\byear{2019})
\end{barticle}
\endbibitem

%%% 7
\bibitem[\protect\citeauthoryear{Nguyen-Duy et~al.}{2023}]{nguyen2023performance}
\begin{barticle}
\bauthor{\bsnm{Nguyen-Duy}, \binits{T.}},
\bauthor{\bsnm{Ngo-Duc}, \binits{T.}},
\bauthor{\bsnm{Desmet}, \binits{Q.}}:
\batitle{Performance evaluation and ranking of cmip6 global climate models over vietnam}.
\bjtitle{Journal of Water and Climate Change}
\bvolume{14}(\bissue{6}),
\bfpage{1831}--\blpage{1846}
(\byear{2023})
\end{barticle}
\endbibitem

%%% 8
\bibitem[\protect\citeauthoryear{Thanh et~al.}{2014}]{thanh2014biascorrection}
\begin{bchapter}
\bauthor{\bsnm{Thanh}, \binits{V.Q.}},
\bauthor{\bsnm{Hoanh}, \binits{C.T.}},
\bauthor{\bsnm{Trung}, \binits{N.H.}},
\bauthor{\bsnm{Tri}, \binits{V.P.D.}}:
\bctitle{A biascorrection method of precipitation data generated by regional climate model}.
In: \bbtitle{International Symposium on Geoinformatics for Spatial Infrastructure Development in Earth and Allied Sciences},
vol. \bseriesno{2014}
(\byear{2014})
\end{bchapter}
\endbibitem

%%% 9
\bibitem[\protect\citeauthoryear{Raftery et~al.}{2005}]{raftery2005using}
\begin{barticle}
\bauthor{\bsnm{Raftery}, \binits{A.E.}},
\bauthor{\bsnm{Gneiting}, \binits{T.}},
\bauthor{\bsnm{Balabdaoui}, \binits{F.}},
\bauthor{\bsnm{Polakowski}, \binits{M.}}:
\batitle{Using bayesian model averaging to calibrate forecast ensembles}.
\bjtitle{Monthly weather review}
\bvolume{133}(\bissue{5}),
\bfpage{1155}--\blpage{1174}
(\byear{2005})
\end{barticle}
\endbibitem

%%% 10
\bibitem[\protect\citeauthoryear{Koenker and Bassett~Jr}{1978}]{koenker1978regression}
\begin{botherref}
\oauthor{\bsnm{Koenker}, \binits{R.}},
\oauthor{\bsnm{Bassett~Jr}, \binits{G.}}:
Regression quantiles.
Econometrica: journal of the Econometric Society,
33--50
(1978)
\end{botherref}
\endbibitem

%%% 11
\bibitem[\protect\citeauthoryear{Gneiting et~al.}{2005}]{gneiting2005calibrated}
\begin{barticle}
\bauthor{\bsnm{Gneiting}, \binits{T.}},
\bauthor{\bsnm{Raftery}, \binits{A.E.}},
\bauthor{\bsnm{Westveld~III}, \binits{A.H.}},
\bauthor{\bsnm{Goldman}, \binits{T.}}:
\batitle{Calibrated probabilistic forecasting using ensemble model output statistics and minimum crps estimation}.
\bjtitle{Monthly weather review}
\bvolume{133}(\bissue{5}),
\bfpage{1098}--\blpage{1118}
(\byear{2005})
\end{barticle}
\endbibitem

%%% 12
\bibitem[\protect\citeauthoryear{Taillardat et~al.}{2016}]{taillardat2016calibrated}
\begin{barticle}
\bauthor{\bsnm{Taillardat}, \binits{M.}},
\bauthor{\bsnm{Mestre}, \binits{O.}},
\bauthor{\bsnm{Zamo}, \binits{M.}},
\bauthor{\bsnm{Naveau}, \binits{P.}}:
\batitle{Calibrated ensemble forecasts using quantile regression forests and ensemble model output statistics}.
\bjtitle{Monthly Weather Review}
\bvolume{144}(\bissue{6}),
\bfpage{2375}--\blpage{2393}
(\byear{2016})
\end{barticle}
\endbibitem

%%% 13
\bibitem[\protect\citeauthoryear{Yeh et~al.}{2016}]{yeh2016matrix}
\begin{bchapter}
\bauthor{\bsnm{Yeh}, \binits{C.-C.M.}},
\bauthor{\bsnm{Zhu}, \binits{Y.}},
\bauthor{\bsnm{Ulanova}, \binits{L.}},
\bauthor{\bsnm{Begum}, \binits{N.}},
\bauthor{\bsnm{Ding}, \binits{Y.}},
\bauthor{\bsnm{Dau}, \binits{H.A.}},
\bauthor{\bsnm{Silva}, \binits{D.F.}},
\bauthor{\bsnm{Mueen}, \binits{A.}},
\bauthor{\bsnm{Keogh}, \binits{E.}}:
\bctitle{Matrix profile i: all pairs similarity joins for time series: a unifying view that includes motifs, discords and shapelets}.
In: \bbtitle{2016 IEEE 16th International Conference on Data Mining (ICDM)},
pp. \bfpage{1317}--\blpage{1322}
(\byear{2016}).
\bcomment{Ieee}
\end{bchapter}
\endbibitem

%%% 14
\bibitem[\protect\citeauthoryear{Zimmerman et~al.}{2019}]{zimmerman2019matrix}
\begin{bchapter}
\bauthor{\bsnm{Zimmerman}, \binits{Z.}},
\bauthor{\bsnm{Kamgar}, \binits{K.}},
\bauthor{\bsnm{Senobari}, \binits{N.S.}},
\bauthor{\bsnm{Crites}, \binits{B.}},
\bauthor{\bsnm{Funning}, \binits{G.}},
\bauthor{\bsnm{Brisk}, \binits{P.}},
\bauthor{\bsnm{Keogh}, \binits{E.}}:
\bctitle{Matrix profile xiv: scaling time series motif discovery with gpus to break a quintillion pairwise comparisons a day and beyond}.
In: \bbtitle{Proceedings of the ACM Symposium on Cloud Computing},
pp. \bfpage{74}--\blpage{86}
(\byear{2019})
\end{bchapter}
\endbibitem

%%% 15
\bibitem[\protect\citeauthoryear{Skamarock et~al.}{2008}]{skamarock2008wrf}
\begin{barticle}
\bauthor{\bsnm{Skamarock}, \binits{W.C.}},
\bauthor{\bsnm{Klemp}, \binits{J.B.}},
\bauthor{\bsnm{Dudhia}, \binits{J.}},
\bauthor{\bsnm{Gill}, \binits{D.O.}},
\bauthor{\bsnm{Barker}, \binits{D.M.}},
\bauthor{\bsnm{Duda}, \binits{M.G.}},
\bauthor{\bsnm{Huang}, \binits{X.-Y.}},
\bauthor{\bsnm{Wang}, \binits{W.}},
\bauthor{\bsnm{Powers}, \binits{J.G.}}:
\batitle{A description of the advanced research wrf version 3}.
\bjtitle{NCAR Technical Note}
\bvolume{475}(\bissue{STR}),
\bfpage{113}
(\byear{2008})
\doiurl{10.5065/D68S4MVH}
\end{barticle}
\endbibitem

%%% 16
\bibitem[\protect\citeauthoryear{Mitsuishi et~al.}{2011}]{Mitsuishi2011}
\begin{barticle}
\bauthor{\bsnm{Mitsuishi}, \binits{S.}},
\bauthor{\bsnm{Ozeki}, \binits{T.}},
\bauthor{\bsnm{Sumi}, \binits{T.}}:
\batitle{Applicability study of a new flood control method using rainfall forecast by wrf (in japanese)}.
\bjtitle{Journal of Japan Society of Hydrology and Water Resources}
\bvolume{24}(\bissue{2}),
\bfpage{110}--\blpage{120}
(\byear{2011})
\end{barticle}
\endbibitem

%%% 17
\bibitem[\protect\citeauthoryear{Thang et~al.}{2022}]{Thang202201}
\begin{barticle}
\bauthor{\bsnm{Thang}, \binits{V.V.}},
\bauthor{\bsnm{Lars}, \binits{R.H.}},
\bauthor{\bsnm{Tien}, \binits{D.D.}},
\bauthor{\bsnm{Kien}, \binits{T.B.}},
\bauthor{\bsnm{Thuc}, \binits{T.D.}}:
\batitle{Assessment of heavy rainfall forecasts over the southern vietnam by using wrf-arw with different physical parameterization schemes}.
\bjtitle{Disaster Advances}
\bvolume{15}(\bissue{2}),
\bfpage{27}--\blpage{44}
(\byear{2022})
\doiurl{10.25303/1502da2744}
\end{barticle}
\endbibitem

%%% 18
\bibitem[\protect\citeauthoryear{Raghavan et~al.}{2016}]{raghavan2016regional}
\begin{barticle}
\bauthor{\bsnm{Raghavan}, \binits{S.}},
\bauthor{\bsnm{Vu}, \binits{M.}},
\bauthor{\bsnm{Liong}, \binits{S.}}:
\batitle{Regional climate simulations over vietnam using the wrf model}.
\bjtitle{Theoretical and Applied Climatology}
\bvolume{126}(\bissue{1}),
\bfpage{161}--\blpage{182}
(\byear{2016})
\end{barticle}
\endbibitem

%%% 19
\bibitem[\protect\citeauthoryear{Wilks}{2011}]{wilks2011statistical}
\begin{bbook}
\bauthor{\bsnm{Wilks}, \binits{D.S.}}:
\bbtitle{Statistical Methods in the Atmospheric Sciences}
vol. \bseriesno{100}.
\bpublisher{Academic press}, \blocation{???}
(\byear{2011})
\end{bbook}
\endbibitem

%%% 20
\bibitem[\protect\citeauthoryear{Ngo-Duc et~al.}{2014}]{ngo2014climate}
\begin{barticle}
\bauthor{\bsnm{Ngo-Duc}, \binits{T.}},
\bauthor{\bsnm{Kieu}, \binits{C.}},
\bauthor{\bsnm{Thatcher}, \binits{M.}},
\bauthor{\bsnm{Nguyen-Le}, \binits{D.}},
\bauthor{\bsnm{Phan-Van}, \binits{T.}}:
\batitle{Climate projections for vietnam based on regional climate models}.
\bjtitle{Climate Research}
\bvolume{60}(\bissue{3}),
\bfpage{199}--\blpage{213}
(\byear{2014})
\end{barticle}
\endbibitem

%%% 21
\bibitem[\protect\citeauthoryear{Ngo-Duc et~al.}{2017}]{ngo2017performance}
\begin{barticle}
\bauthor{\bsnm{Ngo-Duc}, \binits{T.}},
\bauthor{\bsnm{Tangang}, \binits{F.T.}},
\bauthor{\bsnm{Santisirisomboon}, \binits{J.}},
\bauthor{\bsnm{Cruz}, \binits{F.}},
\bauthor{\bsnm{Trinh-Tuan}, \binits{L.}},
\bauthor{\bsnm{Nguyen-Xuan}, \binits{T.}},
\bauthor{\bsnm{Phan-Van}, \binits{T.}},
\bauthor{\bsnm{Juneng}, \binits{L.}},
\bauthor{\bsnm{Narisma}, \binits{G.}},
\bauthor{\bsnm{Singhruck}, \binits{P.}}, \betal:
\batitle{Performance evaluation of regcm4 in simulating extreme rainfall and temperature indices over the cordex-southeast asia region}.
\bjtitle{International Journal of Climatology}
\bvolume{37}(\bissue{3}),
\bfpage{1634}--\blpage{1647}
(\byear{2017})
\end{barticle}
\endbibitem

%%% 22
\bibitem[\protect\citeauthoryear{Gelete}{2023}]{gelete2023application}
\begin{barticle}
\bauthor{\bsnm{Gelete}, \binits{G.}}:
\batitle{Application of hybrid machine learning-based ensemble techniques for rainfall-runoff modeling}.
\bjtitle{Earth Science Informatics}
\bvolume{16}(\bissue{3}),
\bfpage{2475}--\blpage{2495}
(\byear{2023})
\end{barticle}
\endbibitem

%%% 23
\bibitem[\protect\citeauthoryear{Darji}{2019}]{darji2019rainfall}
\begin{botherref}
\oauthor{\bsnm{Darji}, \binits{M.}}:
Rainfall forecasting using neural networks.
International Journal of Research and Analytical Reviews
(2019)
\end{botherref}
\endbibitem

%%% 24
\bibitem[\protect\citeauthoryear{Krishnamurti et~al.}{1999}]{krishnamurti1999improved}
\begin{barticle}
\bauthor{\bsnm{Krishnamurti}, \binits{T.}},
\bauthor{\bsnm{Kishtawal}, \binits{C.M.}},
\bauthor{\bsnm{LaRow}, \binits{T.E.}},
\bauthor{\bsnm{Bachiochi}, \binits{D.R.}},
\bauthor{\bsnm{Zhang}, \binits{Z.}},
\bauthor{\bsnm{Williford}, \binits{C.E.}},
\bauthor{\bsnm{Gadgil}, \binits{S.}},
\bauthor{\bsnm{Surendran}, \binits{S.}}:
\batitle{Improved weather and seasonal climate forecasts from multimodel superensemble}.
\bjtitle{Science}
\bvolume{285}(\bissue{5433}),
\bfpage{1548}--\blpage{1550}
(\byear{1999})
\end{barticle}
\endbibitem

%%% 25
\bibitem[\protect\citeauthoryear{Lakshminarayanan et~al.}{2017}]{lakshminarayanan2017simple}
\begin{botherref}
\oauthor{\bsnm{Lakshminarayanan}, \binits{B.}},
\oauthor{\bsnm{Pritzel}, \binits{A.}},
\oauthor{\bsnm{Blundell}, \binits{C.}}:
Simple and scalable predictive uncertainty estimation using deep ensembles.
Advances in neural information processing systems
\textbf{30}
(2017)
\end{botherref}
\endbibitem

%%% 26
\bibitem[\protect\citeauthoryear{Nilsson et~al.}{2023}]{nilsson2023practical}
\begin{barticle}
\bauthor{\bsnm{Nilsson}, \binits{F.}},
\bauthor{\bsnm{Bouguelia}, \binits{M.-R.}},
\bauthor{\bsnm{R{\"o}gnvaldsson}, \binits{T.}}:
\batitle{Practical joint human-machine exploration of industrial time series using the matrix profile}.
\bjtitle{Data mining and knowledge discovery}
\bvolume{37}(\bissue{1}),
\bfpage{1}--\blpage{38}
(\byear{2023})
\end{barticle}
\endbibitem

%%% 27
\bibitem[\protect\citeauthoryear{De~Paepe et~al.}{2020}]{de2020generalized}
\begin{barticle}
\bauthor{\bsnm{De~Paepe}, \binits{D.}},
\bauthor{\bsnm{Vanden~Hautte}, \binits{S.}},
\bauthor{\bsnm{Steenwinckel}, \binits{B.}},
\bauthor{\bsnm{De~Turck}, \binits{F.}},
\bauthor{\bsnm{Ongenae}, \binits{F.}},
\bauthor{\bsnm{Janssens}, \binits{O.}},
\bauthor{\bsnm{Van~Hoecke}, \binits{S.}}:
\batitle{A generalized matrix profile framework with support for contextual series analysis}.
\bjtitle{Engineering Applications of Artificial Intelligence}
\bvolume{90},
\bfpage{103487}
(\byear{2020})
\end{barticle}
\endbibitem

\end{thebibliography}

\end{document}